\PassOptionsToPackage{table}{xcolor}
\documentclass{prism-uploads/spml_template}

\usepackage[utf8]{inputenc}
\usepackage{amsmath}
\usepackage{amssymb}
\usepackage{enumitem}
\usepackage{fontawesome5}
\makeatletter
\renewcommand{\hyper@natlinkbreak}[2]{#1}
\makeatother
\crefname{section}{Sec.}{Secs.}
\Crefname{section}{Sec.}{Secs.}
\crefname{subsection}{Sec.}{Secs.}
\Crefname{subsection}{Sec.}{Secs.}
\crefname{subsubsection}{Sec.}{Secs.}
\Crefname{subsubsection}{Sec.}{Secs.}
\crefname{appendix}{Appx.}{Appxs.}
\Crefname{appendix}{Appx.}{Appxs.}

\usepackage{etoc}
\usepackage{wrapfig}
\usepackage{needspace}
\usepackage{tikz}
\usetikzlibrary{arrows.meta,fit,positioning}

\definecolor{red}{HTML}{E6194B}
\definecolor{orange}{HTML}{E0925C}
\definecolor{green}{HTML}{3CB44B}
\definecolor{blue}{HTML}{6474CF}
\definecolor{lightblue}{HTML}{B4BEF0}
\definecolor{purple}{HTML}{9169C2}
\definecolor{lightpurple}{HTML}{D3BEEA}
\title{PXRD-Bench: How Close Are We to Determining Crystal Structures from Powder Diffraction?}

\author[1]{Dongyeop Woo}
\author[2]{Nofit Segal}
\author[1]{Nayoung Kim}
\author[1]{Minkyu Kim}
\author[2]{Rafael Gomez-Bombarelli}
\author[1]{Sungsoo Ahn}

\affiliation[1]{Korea Advanced Institute of Science \& Technology (kaist.edu)}
\affiliation[2]{Massachusetts Institute of Technology (mit.edu)}
\metadata[Code]{\href{https://github.com/dongyeop3813/PXRD-Bench.git}{\faGithub\ PXRD-Bench}}
\correspondence{\email{dongyeop@kaist.ac.kr}, \email{sungsoo.ahn@kaist.ac.kr}}

\newcommand{\std}[1]{%
  \,{\scriptsize $\pm$\,#1\phantom{0}}%
}
\newcommand{\model}{Bragglite}
\newcommand{\dataset}{PXRD-Bench}

\begin{document}
\abstract{
Determining crystal structures from powder X-ray diffraction (PXRD) would enable routine characterization of newly synthesized materials, including those absent from reference databases. Generative models offer a promising approach, but existing evaluations remain fragmented and largely assume single-phase samples with known composition. To fill this gap, we introduce \dataset{}, a unified benchmark that extends evaluation to unknown per-element atom counts and recovery of both structures in two-phase mixtures. It provides approximately 4.1 million training structures with simulated PXRD, tests generalization to unseen compositions and structural prototypes, and evaluates recovery from experimental PXRD after excluding structurally matching training entries. To provide a strong reference baseline, we also introduce \model{}, a PXRD-to-structure model developed through a systematic study of PXRD modeling choices. It achieves a 44.2\% Top-1 match rate on experimental RRUFF patterns with known composition, 24.6\% above the strongest prior baseline. Further analysis yields several key findings: some models fail to benefit from experimental PXRD, while generalization to unseen structural prototypes and recovery from two-phase mixtures remain challenging. Together, our benchmark, systematic studies, and \model{} clarify the capabilities and limitations of current PXRD-to-structure models.

}

\maketitle
\addtocontents{toc}{\protect\setcounter{tocdepth}{-1}}

\section{Introduction}
Inferring atomic structures from experimental measurements is central to understanding material properties and guiding materials discovery~\citep{oganov2019structure}. Powder X-ray diffraction (PXRD) is widely used to characterize crystalline materials because it is relatively inexpensive, broadly applicable, and does not require large, high-quality single crystals~\citep{harris2001powder,kaduk2021powder}. Reliable structure determination from PXRD would enable routine characterization of newly synthesized materials, including in high-throughput experiments and self-driving laboratories.

However, averaging diffraction over differently oriented crystallites compresses three-dimensional structural information into a one-dimensional pattern, causing peak overlap and allowing different structures to yield similar patterns~\citep{david2008structure,kaduk2021powder,segal2026loss}. Noise, peak broadening, intensity distortion, and contributions from multiple crystalline phases further complicate interpretation~\citep{altomare2022reliable,hollarek2026opxrd,li2026realpxrdsolver}. Conventional methods combine crystallographic analysis, structure search, and refinement, often requiring substantial expert guidance~\citep{david2008structure,kaduk2021powder}. 

\begin{figure}[t]
  \centering
  \includegraphics[width=\textwidth]{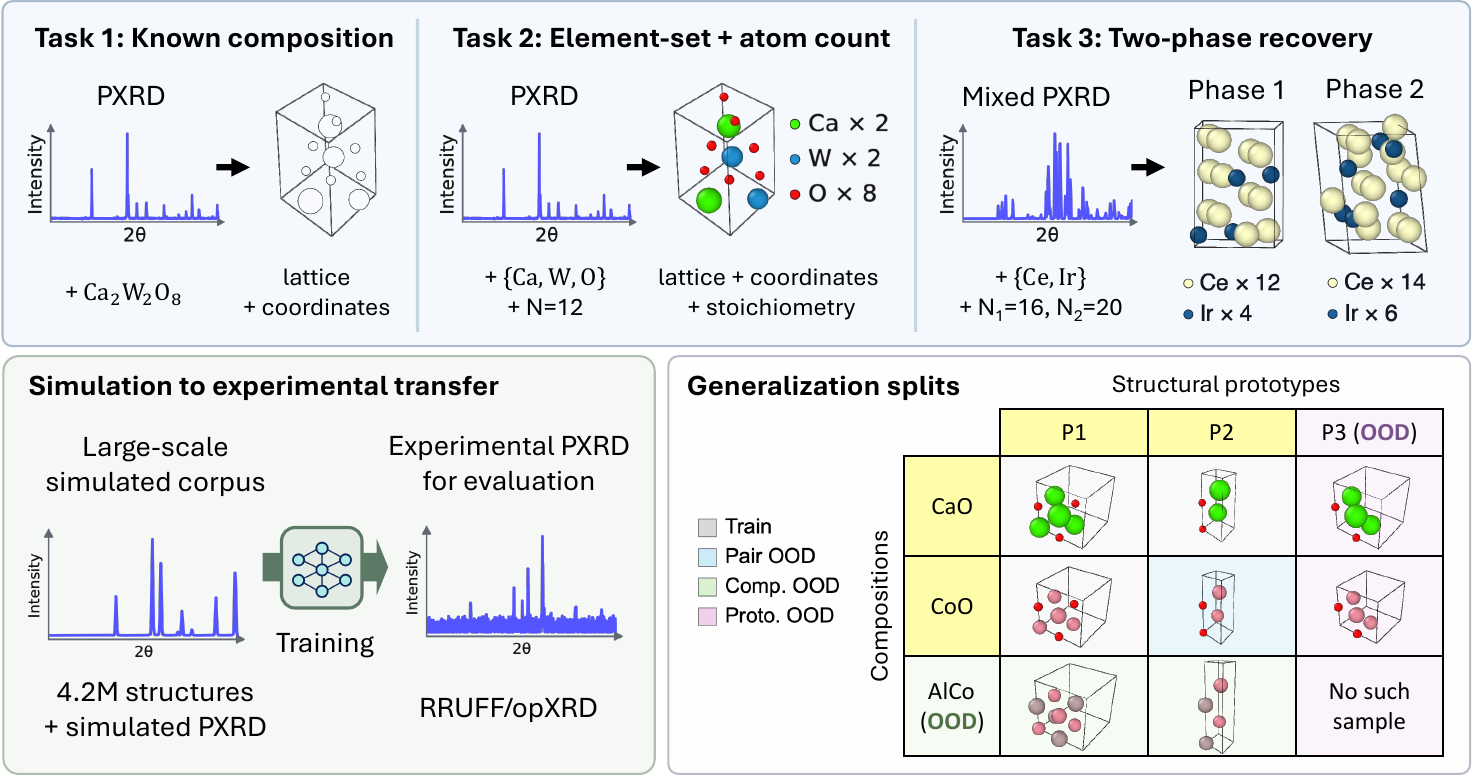}
  \caption{
\textbf{Overview of benchmark tasks and evaluation settings.} Three tasks recover crystal structures from PXRD under known composition, element set, and two-phase conditions.
Models trained on simulated PXRD are evaluated on experimental RRUFF and opXRD patterns.
Generalization splits test unseen composition--prototype pairs, new compositions, and new structural prototypes.
}
  \label{fig:overview-benchmark}
  \vspace{-.1in}
\end{figure}

Generative models offer a promising approach by predicting lattice parameters and atomic positions conditioned on PXRD patterns and available chemical information~\citep{riesel2024crystalyze,li2025pxrdgen,li2026realpxrdsolver,segal2026xrdiff}. Training on known structures paired with simulated PXRD provides abundant data but requires transfer to experimental patterns. Recent work has advanced PXRD encoders and conditional generators~\citep{li2025pxrdgen}, post-generation refinement~\citep{riesel2024crystalyze,li2025pxrdgen}, and representations and augmentation strategies for simulation-to-experiment transfer~\citep{li2026realpxrdsolver,segal2026xrdiff}. 

Despite these advances, the lack of a common benchmark makes it difficult to assess progress and identify remaining challenges. Differences in datasets, chemical inputs, and refinement procedures hinder direct comparison and obscure which design choices drive improvements. Existing evaluations also cover a limited range of settings. Most assume single-phase samples with known composition, leaving recovery with unknown per-element atom counts or overlapping signals from multiple phases underexplored. Furthermore, testing unseen compositions does not establish whether models can recover unfamiliar structural arrangements, since the same structural prototypes may remain represented in training.

\textbf{Contribution.} In this paper, we introduce \dataset{}, a unified benchmark for PXRD-based structure determination. The focus of our new benchmark is twofold:
\begin{itemize}[leftmargin=*]
\item We construct a large-scale corpus of 4.2 million crystal structures paired with simulated PXRD patterns. For experimental evaluation, we curate a benchmark subset from opXRD and adopt the existing RRUFF benchmark, enabling systematic assessment of simulation-to-experiment transfer. We additionally construct simulated two-phase mixtures from thermodynamically plausible phase pairs and evaluate two-phase recovery on experimental GALAXI mixtures.
\item We provide leakage-controlled evaluation protocols that hold out entire compositions, structural prototypes, or composition--prototype pairs. To evaluate models under real-world scenarios, we extend evaluation beyond single-phase samples with known composition to (1) element-set conditioning and (2) two-phase structure recovery. 
\end{itemize}

We further introduce \model{} as a strong baseline for PXRD-based structure determination. \model{} incorporates recent advances in CSP architectures and achieves state-of-the-art performance in our benchmark. Using this architecture, we systematically study how PXRD representations, augmentation strategies, and training-data scale and distribution affect structure recovery and simulation-to-experiment transfer. 

The combination of our new benchmark and systematic studies reveals that current PXRD-to-structure models remain limited in exploiting experimental diffraction and recovering unfamiliar structural prototypes or multiple phases. For example, several existing PXRD-conditioned models underperform DiffCSP on experimental RRUFF patterns, even though DiffCSP receives chemical composition but no PXRD. We observe that data augmentation can improve experimental recovery, but its benefits vary across datasets and may come at the cost of accuracy on simulated patterns. Models also struggle substantially more with unseen structural prototypes than with unseen compositions.

\section{Related Work}
\label{sec:related-work}

\textbf{PXRD-to-structure models.}
Prior work typically formulates structure determination from PXRD as crystal generation conditioned on diffraction patterns and chemical information~\citep{riesel2024crystalyze,li2025pxrdgen,li2026realpxrdsolver,segal2026xrdiff}. Existing methods vary in training data, chemical inputs, and post-generation refinement procedures, complicating direct comparison. Most assume known primitive-cell composition~\citep{riesel2024crystalyze,li2025pxrdgen,li2026realpxrdsolver}, while XRDiff additionally considers element-set conditioning with a known atom count~\citep{segal2026xrdiff}. Prior evaluations also largely focus on in-distribution or composition-level generalization, leaving unseen structural prototypes underexplored. We benchmark models under controlled training data and evaluation settings with held-out  compositions, structural prototypes, and composition--prototype pairs.

\textbf{Transfer to experimental PXRD.}
Because PXRD-to-structure models are typically trained on simulated PXRD, recent work has explored representations designed for experimental robustness and experiment-inspired augmentation for transfer to measured patterns~\citep{riesel2024crystalyze,li2026realpxrdsolver,segal2026xrdiff}. However, these strategies have been studied under different models, training settings, and evaluation protocols, including different experimental PXRD benchmarks, while validation and test sets are often not separated. We compare representation and augmentation choices under a common protocol, use separate experimental validation and test sets, and exclude training structures that match the experimental references.

\textbf{Multiphase PXRD analysis.}
Prior learning-based work on multiphase PXRD has focused on phase identification, i.e., determining which known crystalline phases are present in a mixture, or on separating mixed patterns into phase-specific components. PhaseDifformer~\citep{ito2026phasedifformer} and XDecomposer~\citep{gao2026xdecomposer} learn to decompose mixed PXRD signals, while GALAXI~\citep{tong2026galaxi} addresses multiphase identification from experimental PXRD. These approaches do not directly recover the atomic structures of all constituent phases. We extend multiphase analysis to two-phase structure recovery by generating a crystal structure for each decomposed component. To evaluate transfer to the measured pattern, we use GALAXI's multiphase experimental PXRD dataset.

\section{Benchmark design}
\label{sec:benchmark}


We design \dataset{} around three complementary directions: practically motivated task extensions, simulation-to-experiment transfer, and generalization beyond the training distribution. \Cref{fig:overview-benchmark} summarizes these tasks and evaluation axes.

\subsection{Task formulations}
\label{sec:task-formulation}

We consider three PXRD-based structure determination tasks: single-phase recovery with known composition, single-phase recovery with element-set conditioning~\citep{segal2026xrdiff}, and two-phase recovery. The latter two settings reflect cases where compositional analysis reveals the elements without determining stoichiometry or where a powder sample contains multiple crystalline phases.

We formulate each PXRD inversion task as conditional generation of plausible crystal structures. Let $\mathcal{C}=(\mathbf{A},\mathbf{F},\mathbf{L})$ denote a primitive-cell crystal structure with $N$ atoms, where $\mathbf{A}=(A_1,\ldots,A_N)$ lists their element types, $\mathbf{F}\in[0,1)^{N\times3}$ gives their fractional coordinates, and $\mathbf{L}\in\mathbb{R}^{3\times3}$ is the lattice matrix. For each element $z$, let $c_z=\left|\{i\in\{1,\ldots,N\}:A_i=z\}\right|$ be its primitive-cell atom count. We call $\mathbf{c}=(c_z)_z$ the \emph{composition} and $\mathcal{E}=\{z:c_z>0\}$ the \emph{element set}. Thus $\mathbf{c}$ specifies the multiplicity of every element, whereas $\mathcal{E}$ specifies only which elements are present. We denote the observed PXRD profile by $\mathbf{x}$.

\textbf{Single-phase PXRD tasks.}
In the \emph{known-composition} setting, the model receives the single-phase PXRD profile $\mathbf{x}$ and composition $\mathbf{c}$. The atom types are thereby fixed up to their ordering, and the model generates the remaining structural degrees of freedom as $(\mathbf{F},\mathbf{L})\sim p_{\boldsymbol{\theta}}(\cdot\mid\mathbf{x},\mathbf{c})$.

In the \emph{element-set} setting~\citep{segal2026xrdiff}, the model instead receives $\mathcal{E}$ and the primitive-cell atom count $N$, but not the element counts $\mathbf{c}$. It generates $\mathcal{C}\sim p_{\boldsymbol{\theta}}(\cdot\mid\mathbf{x},\mathcal{E},N)$, requiring it to infer the element multiplicities, fractional coordinates, and lattice.

\textbf{Multiphase PXRD task.}
For a powder sample containing two crystalline phases, we aim to recover both constituent structures from a mixed PXRD profile. We provide mixture-level rather than phase-specific chemical information: the profile $\mathbf{x}_{\mathrm{mix}}$ and the element set $\mathcal{E}_{\mathrm{mix}}=\mathcal{E}_1\cup\mathcal{E}_2$, where $\mathcal{E}_i$ is the element set of phase $\mathcal{C}_i$. We additionally provide the primitive-cell atom counts $N_1$ and $N_2$, but neither phase's composition $\mathbf{c}_i$. The task is to recover
$\{\mathcal{C}_1,\mathcal{C}_2\}\sim
p_{\boldsymbol{\theta}}(\cdot\mid\mathbf{x}_{\mathrm{mix}},
\mathcal{E}_{\mathrm{mix}},N_1,N_2)$.
The model must therefore determine which elements belong to each phase, their multiplicities, and both phases' fractional coordinates and lattices.

\subsection{Single-phase PXRD benchmarks}
For the single-phase benchmarks, we evaluate simulation-to-experiment transfer and generalization to unseen compositions and structural prototypes. We support these evaluations with a large simulated PXRD--structure corpus, controlled held-out splits, and experimental PXRD datasets.

\begin{figure}[!t]
  \centering
  \captionsetup{skip=4pt}
  \captionsetup[subfigure]{skip=2pt}
  \begin{subfigure}[b]{0.64\columnwidth}
    \centering
    \includegraphics[
      width=\linewidth,
      trim=0 30 0 30,
      clip
    ]{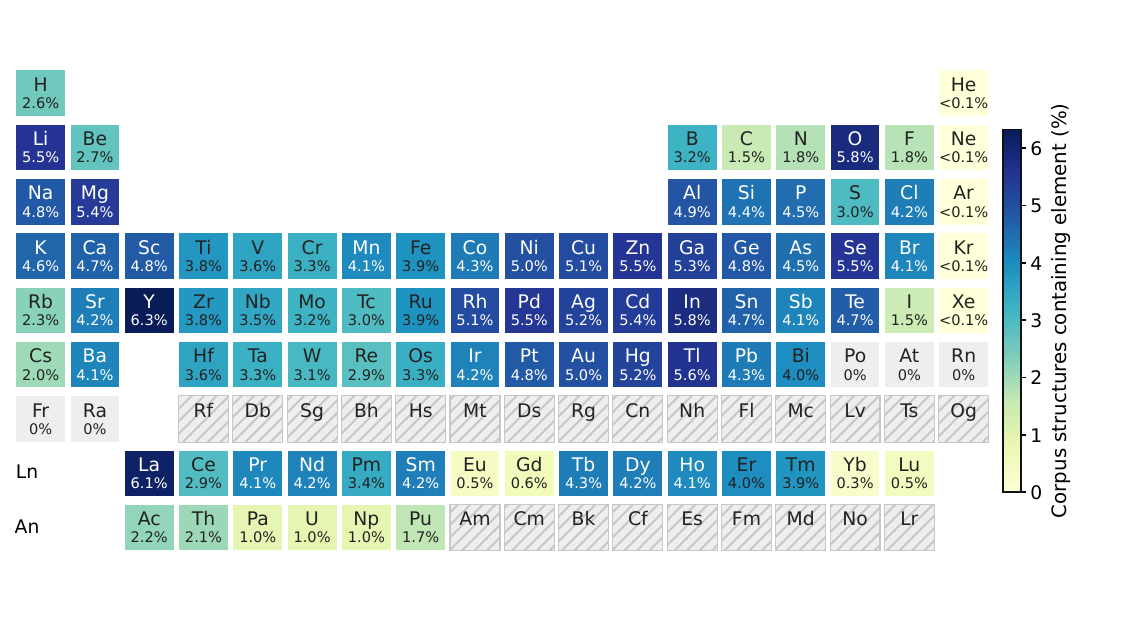}
    \caption{Element occurrence.}
    \label{fig:dataset-elemental-distribution}
  \end{subfigure}
  \hfill
  \begin{subfigure}[b]{0.35\columnwidth}
    \centering
    \includegraphics[
      width=\linewidth,
    ]{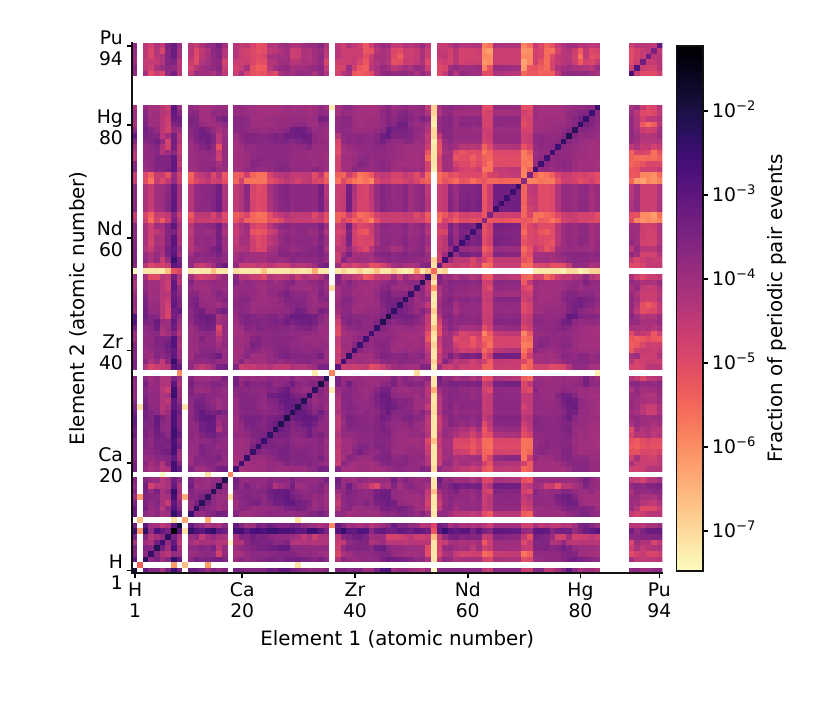}
    \caption{Element-pair frequencies.}
    \label{fig:dataset-pairwise-interactions}
  \end{subfigure}
  \begin{subfigure}[b]{0.32\columnwidth}
    \centering
    \includegraphics[
      width=\linewidth,
    ]{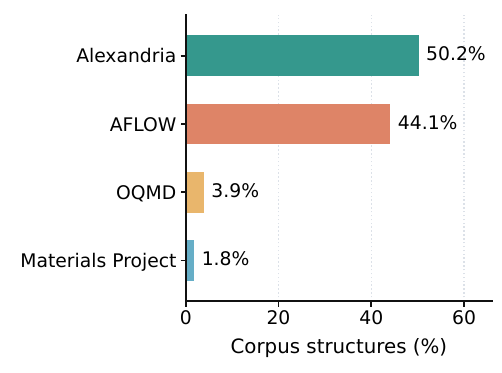}
    \caption{Source provenance.}
    \label{fig:dataset-source-provenance}
  \end{subfigure}
  \hfill
  \begin{subfigure}[b]{0.33\columnwidth}
    \centering
    \includegraphics[
      width=\linewidth,
    ]{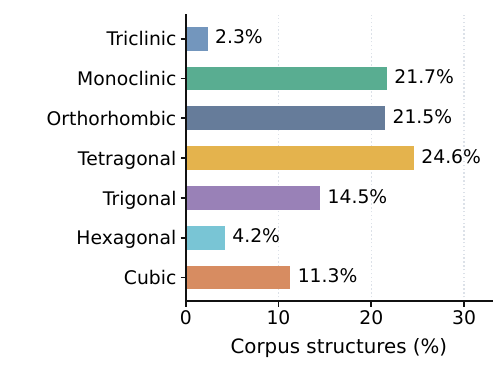}
    \caption{Crystal systems.}
    \label{fig:dataset-crystal-systems}
  \end{subfigure}
  \hfill
  \begin{subfigure}[b]{0.32\columnwidth}
    \centering
    \includegraphics[
      width=\linewidth,
    ]{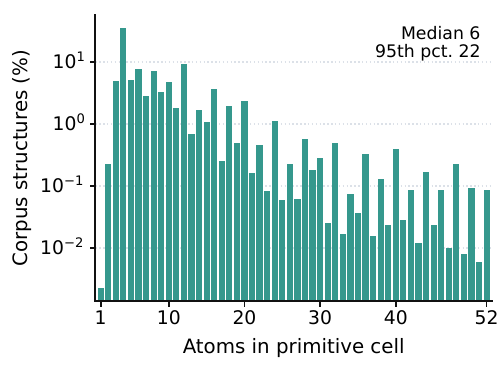}
    \caption{Primitive-cell size.}
    \label{fig:dataset-primitive-cell-size}
  \end{subfigure}
  \caption{\textbf{Coverage of PXRD-Bench's simulated structure/PXRD corpus.}
  (a) Structures containing each element (\%);
  (b) element-pair frequencies normalized by total pair occurrences;
  (c) deduplicated source shares;
  (d) crystal systems; and
  (e) primitive-cell atom counts.}
  \label{fig:dataset-coverage}
  \vspace{-.12in}
\end{figure}
\textbf{Simulated PXRD dataset.}
We construct a corpus of about 4.2 million structures from MP~\citep{jain2013materialsproject,horton2025materialsproject}, Alexandria~\citep{cavignac2026alexandria}, OQMD~\citep{kirklin2015oqmd}, and AFLOW/AFLUX~\citep{curtarolo2012aflow,rose2017aflux}, each paired with a simulated PXRD profile. Preprocessing includes energy-based filtering at 100~meV/atom where energy-above-hull values are available, structural validation, primitive-cell conversion, and cross-source deduplication. The resulting corpus spans all crystal systems and a broad range of elements, with crystal-system, elemental, and element-pair distributions summarized in \cref{fig:dataset-coverage}. Full preprocessing details and additional statistics are provided in \cref{appx:benchmark-simulated-dataset}.

\textbf{Held-out split design.}
We construct one in-distribution (IID) and three out-of-distribution (OOD) splits, all structure-disjoint from training. Composition denotes the reduced composition, i.e., the simplest integer ratio of constituent atom counts, and a structural prototype is defined by the space group and species-anonymized Wyckoff-orbit pattern. We target approximately 20K samples each for validation and testing in every split.

\begin{itemize}[leftmargin=*]
\item \textbf{Pair-OOD}: Specific composition--prototype pairs are held out, while both the composition and prototype remain  represented in training. This tests generalization to unseen combinations of familiar compositions and prototypes.
\item \textbf{Composition-OOD}: Entire compositions are held out from training, testing generalization to unseen chemical compositions.
\item \textbf{Prototype-OOD}: Entire structural prototypes are held out across all compositions, testing generalization to unseen structural prototypes.
\end{itemize}

\textbf{Experimental PXRD datasets.}
To assess simulation-to-real transfer, we use 148 measured \textbf{RRUFF} patterns curated by Crystalyze~\citep{riesel2024crystalyze,lafuente2015rruff}, reserving 48 for validation and 100 for testing. We use the validation set to select PXRD representation and augmentation settings.
Additionally, we curate 123 measured profiles paired with structures from \textbf{opXRD}~\citep{hollarek2026opxrd}. Among samples satisfying our benchmark constraints, we use 40 for validation and 83 for testing. To prevent data leakage, we exclude training structures that structurally match any reference structure in either experimental dataset.

\textbf{Structural matching without refinement.}
Prior work varies in post-generation refinement and structural matching criteria, complicating direct model comparisons. We therefore evaluate unrefined predictions using pymatgen's \texttt{StructureMatcher}~\citep{ong2013pymatgen} with tolerances $(\texttt{ltol}, \texttt{stol}, \texttt{angle\_tol})=(0.2,0.3,5^\circ)$, stricter than the more permissive $(0.3,0.5,10^\circ)$ commonly used in the literature~\citep{jiao2023diffcsp}. These stricter criteria help avoid accepting distorted structures or structures with different space groups~\citep{martirossyan2025structurematches,wei2024cspbench}. Under this protocol, we report top-$k$ match rates for $k\in\{1,20\}$, where a target is considered recovered if at least one of the $k$ candidates matches the reference structure. Metric details are provided in \cref{appx:evaluation-metrics}.

\subsection{Multiphase PXRD benchmark}
For the multiphase benchmark, we construct a simulated two-phase dataset for training and controlled evaluation and additionally evaluate on experimental mixtures.

\textbf{Multiphase PXRD dataset.}
We construct a separate two-phase benchmark based on MP-20. Within binary metallic chemical systems, we use the corresponding phase diagrams to select thermodynamically plausible phase pairs and generate mixtures at multiple relative phase fractions. We retain equilibrium or near-hull assemblages within 50~meV/atom of the convex hull. For each mixture, we combine the constituent PXRD profiles by weighted superposition using Rietveld scale factors, followed by maximum-intensity normalization. We apply this procedure separately within the original MP-20 train, validation, and test splits, yielding 100,798 training, 12,418 validation, and 11,848 test mixtures. Full construction details are provided in \cref{appx:multiphase-dataset}.

\textbf{Experimental multiphase PXRD dataset.}
To assess simulation-to-experiment transfer, we use 21 test samples from GALAXI's impurity-phase experiments~\citep{tong2026galaxi}. We select this subset because its primary--secondary phase mixtures reflect realistic multiphase outcomes of synthesis.

\textbf{Multiphase recovery evaluation.}
We use the same structural matching criterion as in the single-phase benchmark. For each phase, the model generates $k$ candidates, and we report dominant-, minor-, and all-phase top-$k$ match rates for $k\in\{1,20\}$. A sample is counted as an all-phase top-$k$ match only if both constituent phases are recovered within their respective $k$ candidates.

\section{\model{}: A baseline for PXRD-to-structure generation}
\label{sec:crystalite-pxrd}

We introduce \model{}, a PXRD-to-structure reference baseline resulting from our systematic study of PXRD representation, encoding, and augmentation choices. Our model builds on Crystalite~\citep{veljkovic2026crystalite}, a diffusion Transformer with pair-geometry attention biases, and adds a Transformer PXRD encoder to condition on diffraction. We describe the PXRD representation, conditioning, and augmentation used in \model{}; full architecture details are provided in \cref{appx:model-details}.

\subsection{PXRD representation and conditioning}
Following prior peak-based approaches~\citep{li2026realpxrdsolver,segal2026xrdiff}, we represent each PXRD pattern as detected peaks $\mathbf{x}=\{p_i\}_{i=1}^{K}$, with each peak described by its position, intensity, width, and prominence.

\textbf{Peak position representation.}
PXRD peak positions provide information about the lattice parameters. They are conventionally expressed in terms of the diffraction angle $2\theta$, which depends on both crystal structure and X-ray wavelength $\lambda$. We also consider the wavelength-independent interplanar spacing $d=\lambda/(2\sin\theta)$ and scattering-vector magnitude $q=4\pi\sin\theta/\lambda=2\pi/d$. We compare $2\theta$, $d$, and $q$, and use $q$ by default based on our ablation. For $2\theta$ models, we convert each pattern to a canonical wavelength before extracting peaks to remove variation due to measurement wavelength.

\textbf{Peak features.}
Peak intensities depend on the atomic species and their arrangement within the unit cell. Because the absolute intensity scale varies with experimental conditions, we normalize each profile by its maximum intensity and use relative peak intensities. We additionally use peak prominence and width. Prominence measures how strongly a peak stands out from its local background and can help when measured patterns contain background scattering or noise. Peak width captures broadening and can indicate uncertainty in peak localization when peaks are poorly resolved.

\textbf{Peak encoding.}
To provide a richer representation of peak position, we encode $q$ using sinusoidal Fourier features with 16 logarithmically spaced frequencies. We embed the base peak features and Fourier features separately and sum the resulting embeddings to obtain the peak embedding.

\textbf{PXRD conditioning.}
We retain up to 128 highest-intensity peaks and encode them with a Transformer into a profile-level PXRD embedding. For in-context conditioning, we project this embedding to the Crystalite hidden dimension and replicate it across eight context tokens with learnable positional embeddings. The context, atom, and lattice tokens are jointly processed by the Crystalite Transformer.

\subsection{PXRD augmentation}
\begin{figure*}[t]
    \centering
    \includegraphics[width=\textwidth]{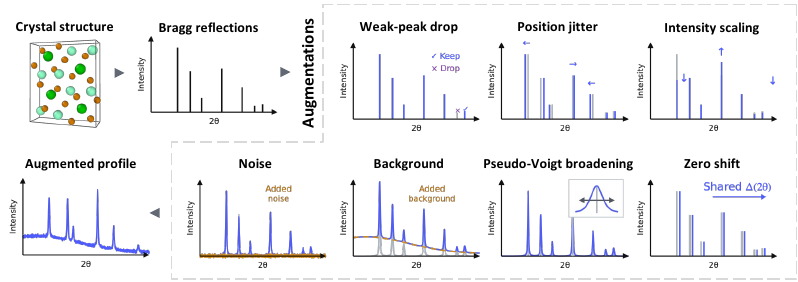}
    \caption{%
        \textbf{Schematic illustration of PXRD profile augmentations.}
        Bragg reflections derived from a crystal structure are perturbed by
        weak-peak removal, position jitter, peak-wise intensity scaling,
        a common angular shift, pseudo-Voigt broadening, background addition,
        and Gaussian noise. Blue denotes perturbed signals and gray denotes
        reference signals.
    }
    \label{fig:pxrd-profile-augmentation}
\vspace{-.2in}
\end{figure*}

To improve transfer from simulated to experimental PXRD, we augment simulated PXRD during training to capture common measurement distortions. Starting from simulated Bragg reflections, we first apply reflection-level augmentations and then construct and augment a dense diffraction profile. \Cref{fig:pxrd-profile-augmentation} illustrates the full pipeline. Exact augmentation definitions, sampling distributions, and hyperparameters are provided in \cref{appx:pxrd-augmentation}. After augmentation, we redetect peaks from the dense profile and use their positions, intensities, widths, and prominences as model input.

\textbf{Reflection-level augmentations.}
We randomly remove weak reflections, independently jitter their positions, and rescale their intensities to model missing peaks, local position errors, and variations in relative peak strength. To model systematic angular calibration errors, we additionally apply a shared zero shift to all reflections.

\textbf{Profile-level augmentations.}
We convert the perturbed Bragg reflections into a dense diffraction profile using pseudo-Voigt peak shapes~\citep{thompson1987profile}. The peak width varies with diffraction angle, with additional randomization to capture variation in peak broadening. We then add a smooth background and Gaussian noise to model background scattering and measurement noise.

\section{Experiments}
\label{sec:experiments}
We first present the single-phase benchmark results (\cref{sec:main-results}), then analyze simulation-to-real transfer in more detail (\cref{sec:sim2real}), and finally evaluate two-phase structure recovery on simulated and experimental mixtures (\cref{sec:multiphase-results}).

\subsection{Single-phase benchmark results}
\label{sec:main-results}
\begin{table}[t]
\caption{
\textbf{Main benchmark results.}
Top-$K$ match rates (\%) on simulated and experimental PXRD datasets.
Results on experimental datasets are reported as mean $\pm$ standard deviation
over five sampling seeds. \textbf{Bold} and \underline{underlined} values indicate the best and second-best
means, respectively, within each dataset and chemical-input setting.
}
\label{tab:benchmark}
\centering
\setlength{\tabcolsep}{5pt}
\renewcommand{\arraystretch}{1.16}
\small
\resizebox{\textwidth}{!}{%
\begin{tabular}{@{\hspace{4pt}}clcccccc@{\hspace{4pt}}}
\toprule
& & \multicolumn{2}{c}{\textbf{Simulated PXRD}}
& \multicolumn{2}{c}{\textbf{RRUFF}}
& \multicolumn{2}{c}{\textbf{opXRD}} \\
\cmidrule(lr){3-4}\cmidrule(lr){5-6}\cmidrule(lr){7-8}
\textbf{Dataset} & \textbf{Model}
& Top-1 & Top-20 & Top-1 & Top-20 & Top-1 & Top-20 \\
\midrule

\multirow{11}{*}{MP-20}
& \multicolumn{7}{l}{\cellcolor{gray!8}\textit{Chemical input: exact stoichiometry}} \\

& PXRDGen
& 54.75 & 77.99
& 30.80\std{3.27} & \underline{57.00}\std{1.41}
& \underline{30.00}\std{1.96} & 36.43\std{0.98} \\

& RealPXRD-Solver
& 56.35 & \underline{80.61}
& 27.80\std{4.76} & 51.20\std{1.10}
& 27.14\std{0.80} & 36.07\std{0.80} \\

& XRDiff
& 55.10 & 78.11
& 17.80\std{0.84} & 39.80\std{0.84}
& 21.07\std{3.87} & 31.43\std{2.04} \\

& Crystalyze
& 29.32 & 48.03
& \phantom{0}5.00\std{1.58} & 14.80\std{1.48}
& 11.07\std{1.49} & 17.86\std{1.79} \\

& DiffCSP (no PXRD)
& 52.17 & 77.28
& 31.60\std{2.70} & \underline{57.00}\std{2.12}
& 23.21\std{1.79} & 36.43\std{0.98} \\

& Crystalite (no PXRD)
& \underline{61.42} & 78.64
& \underline{39.20}\std{2.39} & 52.60\std{1.14}
& 26.43\std{0.80} & \underline{41.43}\std{1.96} \\

& \textbf{\model{} (Ours)}
& \textbf{70.85} & \textbf{82.59}
& \textbf{52.40}\std{0.89} & \textbf{62.60}\std{0.89}
& \textbf{32.14}\std{2.82} & \textbf{42.50}\std{0.80} \\

\cmidrule(lr){2-8}
& \multicolumn{7}{l}{\cellcolor{gray!8}\textit{Chemical input: element set + $N$}} \\

& XRDiff
& \underline{34.29} & \underline{46.13}
& \underline{15.00}\std{1.58} & \underline{22.40}\std{0.55}
& \underline{\phantom{0}9.64}\std{0.98} & \underline{14.64}\std{0.80} \\

& \textbf{\model{} (Ours)}
& \textbf{47.48} & \textbf{71.51}
& \textbf{38.20}\std{2.49} & \textbf{52.80}\std{1.92}
& \textbf{23.57}\std{1.49} & \textbf{28.21}\std{1.49} \\

\midrule

\multirow{10}{*}{\dataset{}}
& \multicolumn{7}{l}{\cellcolor{gray!8}\textit{Chemical input: exact stoichiometry}} \\

& PXRDGen
& \underline{62.50} & 88.90
& \phantom{0}0.00\std{0.00}
& \phantom{0}3.20\std{1.48}
& \phantom{0}0.96\std{0.54}
& \phantom{0}7.71\std{1.83} \\

& RealPXRD-Solver
& 60.45 & \underline{89.50}
& \underline{19.60}\std{2.19}
& 48.00\std{1.22}
& \underline{17.59}\std{1.83}
& 27.47\std{1.79} \\

& XRDiff
& 48.58 & 83.42
& \phantom{0}8.20\std{3.35}
& 37.60\std{3.21}
& \phantom{0}9.16\std{3.14}
& 24.10\std{0.85} \\

& Crystalyze
& 13.92 & 36.26
& \phantom{0}0.60\std{0.55}
& \phantom{0}2.60\std{1.34}
& \phantom{0}0.72\std{0.66}
& \phantom{0}3.13\std{1.08} \\

& Crystalite (no PXRD)
& 30.54 & 75.49
& 16.20\std{4.09}
& \underline{57.80}\std{2.68}
& 11.08\std{1.57}
& \underline{31.57}\std{3.55} \\

& \textbf{\model{} (Ours)}
& \textbf{76.48} & \textbf{93.81}
& \textbf{44.20}\std{2.49}
& \textbf{69.20}\std{2.05}
& \textbf{23.61}\std{1.08}
& \textbf{38.80}\std{1.32} \\

\cmidrule(lr){2-8}
& \multicolumn{7}{l}{\cellcolor{gray!8}\textit{Chemical input: element set + $N$}} \\

& XRDiff
& \underline{22.45} & \underline{49.78}
& \phantom{0}\underline{2.20}\std{1.48}
& \underline{13.80}\std{1.48}
& \phantom{0}\underline{6.51}\std{1.83}
& \underline{11.57}\std{1.08} \\

& \textbf{\model{} (Ours)}
& \textbf{56.53} & \textbf{81.59}
& \textbf{36.00}\std{2.12}
& \textbf{58.80}\std{1.30}
& \textbf{15.18}\std{0.66}
& \textbf{24.58}\std{2.02} \\
\bottomrule
\end{tabular}%
}
\vspace{-.12in}
\end{table}
We evaluate single-phase structure recovery under the known-composition and element-set settings defined in \cref{sec:task-formulation}. We benchmark models trained on \dataset{} and additionally report MP-20~\citep{jain2013materialsproject,jiao2023diffcsp} results for reference. We compare \model{} with Crystalyze~\citep{riesel2024crystalyze}, PXRDGen~\citep{li2025pxrdgen}, RealPXRD-Solver~\citep{li2026realpxrdsolver}, and XRDiff~\citep{segal2026xrdiff} under controlled training data and evaluation settings, and also include DiffCSP~\citep{jiao2023diffcsp} and Crystalite~\citep{veljkovic2026crystalite} as composition-only baselines without PXRD input. \Cref{tab:benchmark} reports match rates on the simulated in-distribution test sets and on experimental RRUFF and opXRD patterns. For opXRD, MP-20-trained models are evaluated only on $\leq20$-atom structures, while \dataset{}-trained models use the full set. Detailed baseline configurations and experimental settings appear in \cref{appx:experiment-details}.

\textbf{\model{} outperforms baselines across all settings.} 
In the known composition setting, \model{} achieves higher match rates than existing baselines on both simulated and experimental PXRD across the two training datasets. It also outperforms XRDiff under element-set conditioning.



\textbf{PXRD-conditioned models can underperform CSP baselines.}
On MP-20 and RRUFF, Crystalite achieves higher Top-1 match rates than existing PXRD-conditioned baselines. This suggests that a strong composition prior alone often recovers the reference structure when polymorph ambiguity is limited. In contrast, Crystalite performs substantially worse on the polymorph-rich \dataset{}, where PXRD is needed to distinguish structures sharing the same composition.

\textbf{Existing baselines struggle with experimental PXRD.}
To isolate the contribution of PXRD conditioning from backbone differences, we train a composition-only DiffCSP model with the same backbone and model size as the existing baselines. On RRUFF, this CSP-only model matches or outperforms its PXRD-conditioned counterparts, suggesting that PXRD representations optimized on simulated data do not transfer robustly to experimental measurements.

\begin{table}[t]
\caption{
\textbf{Out-of-distribution generalization.}
Top-$K$ match rates (\%) on the OOD simulated PXRD splits of \dataset{}, grouped by chemical input. \textbf{Bold} indicates the best result within each setting.
}
\label{tab:generalization}
\centering
\begingroup
\small
\setlength{\tabcolsep}{5pt}
\renewcommand{\arraystretch}{1.15}
\resizebox{0.75\linewidth}{!}{%
\begin{tabular}{lcccccc}
\toprule
& \multicolumn{2}{c}{\textbf{Pair OOD}}
& \multicolumn{2}{c}{\textbf{Prototype OOD}}
& \multicolumn{2}{c}{\textbf{Composition OOD}} \\
\cmidrule(lr){2-3}\cmidrule(lr){4-5}\cmidrule(lr){6-7}
\textbf{Model} & Top-1 & Top-20 & Top-1 & Top-20 & Top-1 & Top-20 \\
\midrule
\multicolumn{7}{l}{\cellcolor{gray!8}\textit{Chemical input: exact stoichiometry}} \\
PXRDGen
& 62.40 & 89.98
& 7.55 & 33.97
& 63.80 & 89.61 \\

RealPXRD-Solver
& 58.94 & 88.88
& 3.31 & 17.52
& 62.90 & 90.66 \\

XRDiff
& 47.48 & 83.87
& 2.36 & 18.51
& 50.13 & 84.16 \\

Crystalyze
& 16.06 & 39.53
& 0.58 & 3.19
& 14.00 & 36.91 \\

Crystalite (no PXRD)
& 16.81 & 69.75
& 1.39 & 11.12
& 27.30 & 73.32 \\

\textbf{\model{} (Ours)}
& \textbf{75.26} & \textbf{94.18}
& \textbf{12.37} & \textbf{35.89}
& \textbf{77.27} & \textbf{94.55} \\
\midrule
\multicolumn{7}{l}{\cellcolor{gray!8}\textit{Chemical input: element set + $N$}} \\

XRDiff
& 21.89 & 49.57
& 0.59 & 4.49
& 24.02 & 51.39 \\

\textbf{\model{} (Ours)}
& \textbf{54.46} & \textbf{80.01}
& \textbf{4.08} & \textbf{13.86}
& \textbf{57.27} & \textbf{82.13} \\
\bottomrule
\end{tabular}
}
\par
\endgroup
\noindent
\vspace{-.12in}
\end{table}

\textbf{Generalization beyond training distribution.}
We evaluate models on held-out splits with unseen composition--prototype pairings, chemical compositions, and structural prototypes. As shown in \cref{tab:generalization}, performance remains strong on unseen pairings and compositions but drops substantially on unseen structural prototypes. This indicates that generalization to unseen pairings and compositions is relatively tractable when corresponding structural prototypes are represented during training, whereas generalization to unseen structural prototypes remains challenging. To assess whether recovery of unseen structural prototypes is driven solely by the learned crystal structure prior, we compare \model{} with its composition-only counterpart, Crystalite. Despite low absolute match rates, \model{} substantially outperforms the CSP baseline, indicating that PXRD provides structure-specific 
information beyond the composition prior even for unseen structural prototypes.

\begin{wrapfigure}{r}{0.31\linewidth}
  \centering
  \includegraphics[width=\linewidth]{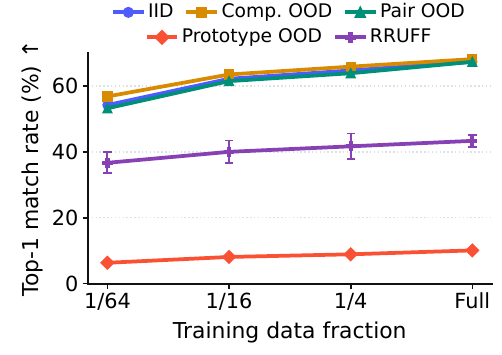}
  \captionsetup{
    skip=2pt
  }
  \caption{
    \textbf{Data scaling.}
    Performance vs.\ training data fraction.
  }
  \label{fig:data-scaling}
\end{wrapfigure}
\textbf{Data scaling experiments.}
We train \model{} on progressively larger subsets of \dataset{} to study the effect of simulated training scale. As shown in \cref{fig:data-scaling}, Top-1 match rates improve across all simulated splits and on experimental RRUFF as data scale increases. The gains diminish at larger scales but remain consistent across simulated and experimental evaluations.

\subsection{Understanding sim-to-real transfer}
\label{sec:sim2real}

We study how PXRD representation and augmentation affect simulation-to-real transfer while fixing the model architecture and training configuration. Unless otherwise stated, we train \model{} on MP-20 for 200K optimizer steps and vary only the PXRD representation or augmentation strategy. We focus here on the effects of PXRD augmentation and training distribution, while additional ablations of PXRD representations are provided in \cref{appx:pxrd-representation-study}.

\begin{figure}[!t]
  \centering
  \captionsetup{skip=6pt}
  \includegraphics[
    width=\textwidth,
    keepaspectratio
  ]{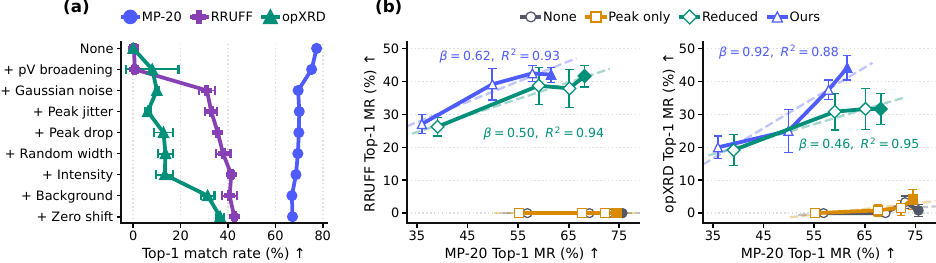}
  \caption{
    \textbf{Analysis of PXRD augmentation and simulation-to-experiment transfer.}
    (a) Additive ablation of the PXRD augmentation components.
    (b) Relationship between simulated MP-20 and experimental validation
    performance on RRUFF and opXRD. Filled markers denote the final training checkpoint; solid lines connect results across training checkpoints.
  }
  \label{fig:augmentation-analysis}
  \phantomsubcaption
  \label{fig:augmentation-ablation}
  \phantomsubcaption
  \label{fig:profile-augmentation-correlation}
\end{figure}

\textbf{Ablation on PXRD augmentation.}
We progressively add profile-augmentation components to assess their contributions. As shown in \cref{fig:augmentation-ablation}, training without augmentation achieves high Top-1 accuracy on simulated PXRD but transfers poorly to experimental RRUFF. Adding augmentations consistently improves RRUFF performance while generally reducing simulated accuracy, indicating a trade-off between simulation performance and experimental robustness. Furthermore, individual components vary substantially in effectiveness across experimental datasets, indicating that augmentations effective for one dataset may not transfer equally well to others. Developing sim-to-real augmentations that generalize across diverse experimental artifacts and conditions therefore remains important.

\textbf{Correlation between simulated and experimental performance.}
We examine whether simulated and experimental performance follow similar trends across augmentation strategies. We compare no augmentation, peak-only augmentation, and the full pipeline with reduced or proposed augmentation magnitudes. During 200K-step training, we evaluate Top-1 validation match rates on MP-20, RRUFF, and opXRD every 50K steps and fit linear trends across four checkpoints. As shown in \cref{fig:profile-augmentation-correlation}, no augmentation and peak-only augmentation yield little experimental improvement as simulated performance increases, whereas both full-augmentation settings show positive trends, with the proposed magnitudes yielding larger experimental gains.


\textbf{Effect of training distribution.}
Despite its smaller size, MP-20 transfers better to RRUFF than \dataset{}.
RRUFF is primarily composed of minerals and is enriched in elements such as Si
and O, which are more frequently represented in MP-20. As shown in \cref{tab:training-distribution}, oversampling training structures containing these mineral-associated elements further improves RRUFF
performance, suggesting that distribution-aware fine-tuning on \dataset{}
toward the expected target domain may further improve performance on downstream
experimental PXRD.

\begin{table*}[t]
\centering
\captionsetup{
    skip=2pt,
}

\begin{minipage}[t]{0.46\textwidth}
\vspace{0pt}
\captionof{table}{
\textbf{Effect of Si/O oversampling.}
Validation match rates (\%) on RRUFF.
}
\label{tab:training-distribution}

\centering
\begingroup
\small
\renewcommand{\arraystretch}{0.92}
\setlength{\tabcolsep}{3.5pt}

\begin{tabular}{@{}lcc@{}}
\toprule
& \multicolumn{2}{c}{\textbf{RRUFF}} \\
\cmidrule(lr){2-3}
\textbf{Training setup}
& Top-1
& Top-20 \\
\midrule
MP-20
& 48.33\std{1.74}
& 58.33\std{2.08} \\
PXRD-Bench
& 43.75\std{2.95}
& 70.42\std{1.74} \\
\hspace{0.3em}+ oversampling
& 54.58\std{3.42}
& 74.17\std{1.14} \\
\bottomrule
\end{tabular}

\endgroup
\end{minipage}
\hfill
\begin{minipage}[t]{0.50\textwidth}
\vspace{0pt}
\captionof{table}{
\textbf{Multiphase recovery.}
All-phase Top-$K$ match rates (\%) on MP-20 and GALAXI.
}
\label{tab:multiphase-recovery}

\centering
\begingroup
\small
\renewcommand{\arraystretch}{0.92}
\setlength{\tabcolsep}{2.5pt}

\begin{tabular}{@{}lcccc@{}}
\toprule
& \multicolumn{2}{c}{\textbf{MP-20}}
& \multicolumn{2}{c}{\textbf{GALAXI}} \\
\cmidrule(lr){2-3}
\cmidrule(lr){4-5}
\textbf{Model}
& Top-1 & Top-20
& Top-1 & Top-20 \\
\midrule
MatterGen
& 2.13 & \textbf{30.67}
& 0.00 & 42.86 \\
XRDiff
& 6.76 & 14.64
& 0.00 & 9.52 \\
\textbf{\model{}}
& \textbf{13.26} & 25.34
& \textbf{19.05} & \textbf{47.62} \\
\bottomrule
\end{tabular}

\endgroup
\end{minipage}

\vspace{-0.12in}
\end{table*}

\subsection{Two-phase recovery} 
\label{sec:multiphase-results}

We evaluate two-phase structure recovery on the simulated MP-20 benchmark
and experimental PXRD mixtures from GALAXI~\citep{tong2026galaxi}. Each
mixed profile is first decomposed into phase-specific profiles using
XDecomposer~\citep{gao2026xdecomposer}, after which a structure generator
is applied independently to each profile. Starting from the
element-set-conditioned \model{}, we fine-tune it so that each phase-specific
generation uses the shared mixture-level element set.

We use MatterGen~\citep{zeni2025generative} and XRDiff as per-phase candidate generators, but treat
them as reference methods rather than strictly controlled baselines:
MatterGen is not leakage-controlled against our evaluation sets, while XRDiff
receives phase-specific element sets. As shown in
\cref{tab:multiphase-recovery}, \model{} outperforms the reference methods
in most settings. Its substantially higher Top-1 match rate than the
PXRD-free MatterGen reference further supports the value of diffraction
conditioning.
\section{Conclusion}

We introduce \dataset{}, a unified benchmark for PXRD-based structure determination spanning simulation-to-experiment transfer, out-of-distribution generalization, element-set conditioning, and two-phase recovery. We also develop \model{}, a strong PXRD-conditioned baseline, and use it to systematically study PXRD representations and augmentation strategies. Our experiments show that \model{} substantially improves experimental structure recovery over prior methods, while revealing that robust use of experimental PXRD, generalization to unseen structural prototypes, and recovery of multiple phases remain challenging. These results establish \dataset{} as a controlled framework for measuring progress toward practical PXRD-based structure determination and identify key limitations that future methods must address.

\subsection*{AI use statement}
We used OpenAI's ChatGPT and Codex to support implementation across the research workflow, including data processing, model development, training, evaluation, and interpretation of experimental results. These tools were also used to assist with plotting and analysis code, figure and table preparation, and drafting, revising, and formatting the manuscript. Their use was guided by the authors' research ideas, experimental goals, and intended analyses. Generative AI did not independently determine the research questions, core methodology, or final scientific conclusions. The authors reviewed the resulting analyses, figures, tables, and manuscript text and take full responsibility for the final results, claims, and presentation of this work.

\subsection*{Reproducibility statement}
We provide the details needed to reproduce the benchmark construction and experiments throughout the main text and appendix. The benchmark task definitions, dataset construction and splits, leakage controls, and evaluation metrics are described in \cref{sec:benchmark} and \cref{appx:benchmark-detail}. Model architecture and PXRD preprocessing and augmentation are specified in \cref{sec:crystalite-pxrd} and \cref{appx:model-details}. \cref{appx:experiment-details} reports the training objectives, optimization hyperparameters, sampling procedure, checkpoint-selection protocol, baseline configurations, random seeds, and compute resources used for the reported experiments. We additionally provide the implementation code and experiment configurations in the supplementary material.

\bibliographystyle{plainnat}
\bibliography{references}

\newpage
\appendix
\raggedbottom

\crefalias{section}{appendix}
\crefalias{subsection}{appendix}
\crefalias{subsubsection}{appendix}

\etocdepthtag.toc{appendix}

\newpage
\etocsettagdepth{appendix}{subsection}
\etocsettagdepth{default}{none}
\renewcommand{\contentsname}{Appendix}
\begingroup
\renewcommand{\etocaftertitlehook}{\setlength{\parskip}{0.35\baselineskip}}
\tableofcontents
\endgroup

\newpage
\section{Benchmark details}
\label{appx:benchmark-detail}

This section provides additional details on dataset construction, curation, and evaluation metrics for our benchmark.

\subsection{Large-scale simulated PXRD dataset}
\label{appx:benchmark-simulated-dataset}

\subsubsection{Dataset construction}
We construct the simulated benchmark through source-specific selection,
structural validation, composition-bucket filtering, cross-source
deduplication, and final benchmark curation.

\textbf{Source-specific selection.}
The frozen snapshots of MP~\citep{jain2013materialsproject,horton2025materialsproject}, OQMD~\citep{kirklin2015oqmd}, Alexandria~\citep{cavignac2026alexandria}, and AFLOW/AFLUX~\citep{curtarolo2012aflow,rose2017aflux} contain approximately 9.78 million input records in total. This aggregate should not be interpreted as a count of unique structures: it includes repeated AFLOW API records, and the source-specific counts refer to different processing stages. We apply energy-based selection to MP, OQMD, and Alexandria, but no energy filter to AFLOW because the AFLOW snapshot does not store energy-above-hull values for assessing thermodynamic stability. This source-specific selection retains 5,634,620 candidate records. \Cref{tab:dataset-source-selection} summarizes the snapshots and selection criteria.

\textbf{Structural validation.}
We next validate the selected structures and convert them to primitive cells, retaining only three-dimensionally periodic structures with fully occupied, ordered sites. We exclude structures with invalid cell volumes (non-finite or $\leq 10^{-6}$~\AA$^3$), no atomic sites, non-finite lattice parameters or atomic coordinates, or elements with atomic numbers greater than 94. At this stage, we also discard structures with more than 208 atoms in the primitive cell.

\textbf{Composition-bucket filtering and deduplication.}
Before deduplication, we group candidate records by reduced composition and discard every bucket containing more than 128 records in its entirety. We then deduplicate the remaining structures across all sources using sequential, representative-centered, species-aware structural matching with $(\mathrm{ltol},\mathrm{stol},\mathrm{angle\_tol})=(0.2,0.3,5^\circ)$. Representatives are selected using a fixed source priority of MP, OQMD, Alexandria, then AFLOW. This priority is fixed for reproducibility and does not rank source quality. We also retain the provenance of matching records from other sources.

\textbf{Atom-count restriction and chemical curation.}
For the benchmark used in our experiments, we further restrict the corpus to structures with at most 52 atoms in the primitive cell and apply chemical curation to focus the dataset on extended inorganic crystalline materials.

We screen structures using periodic bond connectivity to identify discrete molecular units and extended networks. We exclude molecular crystals composed of finite, neutral molecular units, including both organic and inorganic molecules. We additionally exclude structures with evidence of organic components, including organic ions, polymeric organic structures, metal--organic coordination networks, COF-like networks, and organic guests in inorganic hosts. This broader screening does not require definitive classification as a salt, MOF, or COF, nor confirmed porosity, and conservatively excludes ambiguous candidates with evidence of organic components. This curation is an operational exclusion policy rather than a guarantee that the remaining corpus is entirely free of molecular or organic structures.

After all filtering and split-consistency checks described below, the resulting benchmark contains 4,262,691 structures, including an effective training set of 4,107,627 structures. Alexandria and AFLOW account for the largest shares of representative structures in the final corpus (\cref{fig:dataset-source-provenance}).

For each retained structure, we simulate a PXRD profile using Cu K$\alpha$ radiation over $2\theta\in[5^\circ,90^\circ]$.

\begin{table}[h]
  \centering
  \caption{
Frozen source snapshots, selection criteria, and record counts before and
after dataset curation. Candidate counts are measured after structural
validation and before composition-bucket filtering, cross-source
deduplication, and the 52-atom restriction. Final counts are measured after
the 52-atom restriction, chemical curation, OOD-consistency filtering, and
benchmark-overlap exclusion, totaling 4,262,691 structures, including
4,107,627 training structures. Each retained structure is attributed to the
source of its representative record, selected during deduplication using the
fixed priority MP, OQMD, Alexandria, then AFLOW. The final counts therefore
reflect representative provenance rather than independent source
contributions or source quality. $E_{\mathrm{hull}}$ denotes energy above
hull; OQMD uses its signed stability field, including negative values. All
energy values are in eV/atom.
  }
  \label{tab:dataset-source-selection}
  \footnotesize
  \setlength{\tabcolsep}{3pt}
  \renewcommand{\arraystretch}{1.15}
  \begin{tabular*}{\linewidth}{@{\extracolsep{\fill}}p{0.15\linewidth}p{0.26\linewidth}p{0.27\linewidth}rr@{}}
    \toprule
    Source & Snapshot & Selection criteria
      & \multicolumn{2}{c}{Record counts} \\
    \cmidrule(l){4-5}
      & & & Candidates & Final corpus \\
    \midrule
    \raggedright Materials Project
      & \raggedright 2026-04-13
      & \raggedright $0 \leq E_{\mathrm{hull}} \leq 0.10$
      & 100,384 & 77,493 \\
    \addlinespace[3pt]
    OQMD
      & \raggedright v1.5 (2021-10)
      & \raggedright Signed stability $\leq 0.10$;
        negative values retained
      & 219,945 & 166,102 \\
    \addlinespace[3pt]
    Alexandria
      & \raggedright PBE 3D, 2025-07-02
      & \raggedright $0 \leq E_{\mathrm{hull}} \leq 0.10$
      & 2,331,796 & 2,139,337 \\
    \addlinespace[3pt]
    AFLOW
      & \raggedright AFLUX, 2026-08-16
      & \raggedright No energy filter
      & 2,981,679 & 1,879,759 \\
    \midrule
    \multicolumn{3}{@{}l}{Total}
      & 5,633,804 & 4,262,691 \\
    \bottomrule
  \end{tabular*}
\end{table}

\subsubsection{Dataset splits and overlap audits}
We use a sequential holdout procedure to partition the corpus into a
training set and validation/test sets for four evaluation settings: IID,
prototype-OOD, composition-OOD, and composition--prototype-pair-OOD
(pair-OOD). We construct the three OOD holdouts first, followed by the IID
holdouts.

We define a structural prototype by its space group, primitive-cell site
count, and species-anonymized per-species multiset of Wyckoff orbits, using
a symmetry tolerance of $\mathrm{symprec}=0.1$~\AA. To prevent structurally
similar examples from crossing split boundaries, we additionally group
structures using species-aware structural matching with
$(\mathrm{ltol},\mathrm{stol},\mathrm{angle\_tol})
=(0.3,0.5,10^\circ)$. Structural similarity groups are defined as connected
components of the resulting matching graph and are kept intact during
splitting.

First, we construct the prototype-OOD set. A candidate prototype is eligible
for holdout only if every associated reduced composition remains represented
in the training pool by another prototype. We hold out all structures
belonging to each selected prototype, targeting approximately 40,000
structures in total.

Second, we construct the composition-OOD set, excluding reduced compositions
associated with the prototype-OOD set from consideration. We hold out all
structures with a candidate composition only if every associated prototype
remains represented in the training pool by another composition, again
targeting approximately 40,000 structures.

Third, we construct the pair-OOD set from structures not assigned to either
preceding OOD set. We hold out complete composition--prototype pairs only if
the composition and prototype each remain represented in the training pool
through other pairs. A pair is eligible only if it can be held out without
splitting any structural similarity group.

We divide each OOD set into validation and test subsets, targeting
approximately 20,000 structures per subset. The defining holdout
units---prototypes, reduced compositions, or composition--prototype
pairs---are kept intact, so that no unit is shared between the corresponding
validation and test subsets. After constructing the OOD splits, we randomly hold out structural similarity groups from the remaining pool to form IID validation and test sets of approximately 20,000 structures each. We refer to these random structural-group holdouts as IID splits because they do not explicitly enforce novelty in composition or structural prototype. Consequently, rare compositions or prototypes may occur only in validation or test under random group assignment.

Subsequent filtering can remove the remaining training examples needed to
satisfy an OOD holdout condition. We therefore re-evaluate the OOD
constraints after the IID holdouts and again after the atom-count,
chemical-curation, and benchmark-overlap exclusions described above and
below. Any validation or test structures that no longer satisfy their
defining OOD condition with respect to the final effective training set are
removed.

To prevent training overlap with the external benchmarks, we exclude
training structures that match reference structures in RRUFF-148 or
opXRD-123. Matching requires identical element species and uses
primitive-cell conversion and lattice scaling, without supercell or subset
matching. For this exclusion, we use more permissive structural matching
tolerances than those used for the match-rate metric:
$(\mathrm{ltol},\mathrm{stol},\mathrm{angle\_tol})
=(0.3,0.5,10^\circ)$.

Audits of the finalized splits confirm that no structural similarity group
is shared across splits and that no training structure matches an
RRUFF-148 or opXRD-123 reference after exclusion. These checks separately
address internal split leakage and training--benchmark structural overlap.
\Cref{fig:dataset-split-sizes} reports the finalized split sizes.

\subsubsection{Additional dataset coverage statistics}

\begin{figure}[p]
  \centering
  \begin{minipage}{\linewidth}
  \centering
  \makebox[\linewidth][c]{\hspace*{-0.24\textwidth}\includegraphics[width=0.62\textwidth]{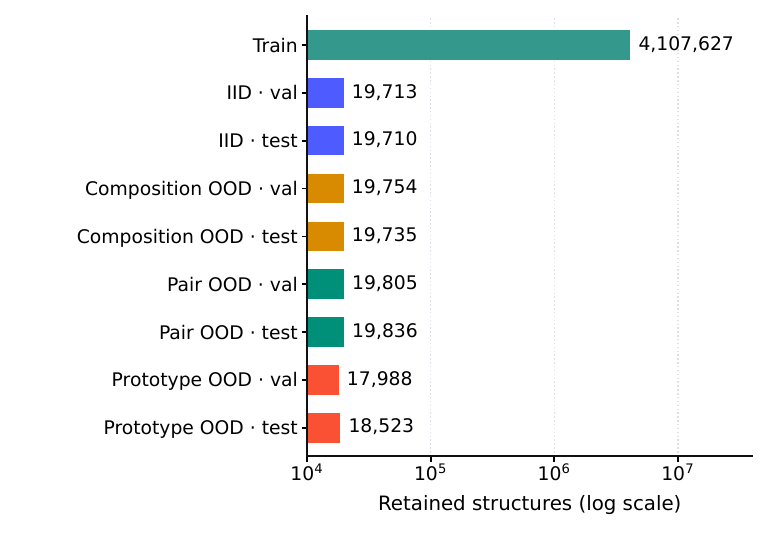}}
  \caption{Finalized split sizes of the large-scale simulated corpus.
  The plot reports the training-set size and validation/test sizes for IID,
  pair-OOD, prototype-OOD, and composition-OOD evaluations on a logarithmic
  horizontal axis. No structural similarity group is shared across splits, and training--benchmark
  structural overlap with RRUFF-148 and opXRD-123 is zero after exclusion.}
  \label{fig:dataset-split-sizes}
  \end{minipage}
  \medskip
  \nextfloat
  \begin{minipage}{\linewidth}
  \centering
  \begin{subfigure}[t]{0.47\linewidth}
    \centering
    \includegraphics[width=\linewidth]{resource/image/primitive_cell_size.pdf}
    \caption{Primitive-cell size.}
    \label{fig:dataset-atom-count}
  \end{subfigure}%
  \begin{subfigure}[t]{0.47\linewidth}
    \centering
    \includegraphics[width=\linewidth]{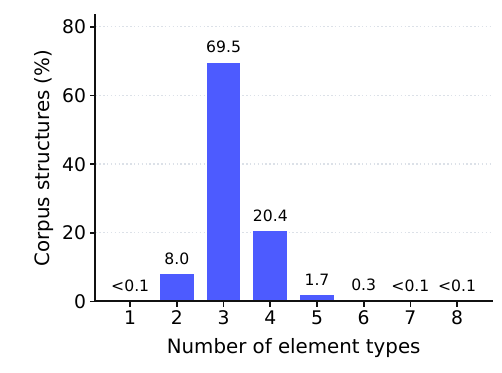}
    \caption{Composition arity.}
    \label{fig:dataset-composition-arity}
  \end{subfigure}

  \begin{subfigure}[t]{0.47\linewidth}
    \centering
    \includegraphics[width=\linewidth]{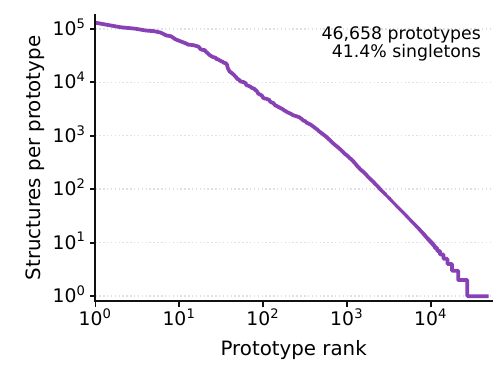}
    \caption{Prototype frequencies.}
    \label{fig:dataset-prototype-frequency}
  \end{subfigure}%
  \begin{subfigure}[t]{0.47\linewidth}
    \centering
    \includegraphics[width=\linewidth]{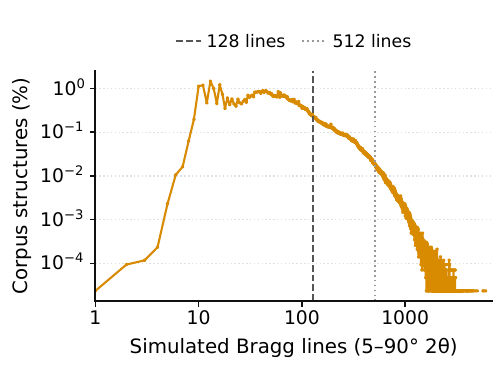}
    \caption{Simulated Bragg line counts.}
    \label{fig:dataset-peak-complexity}
  \end{subfigure}
  \caption{Structural and diffraction-profile distributions of the large-scale
  simulated corpus. (a) Primitive-cell atom counts. (b) Number of distinct
  elements per structure. (c) Structures per prototype, ordered by decreasing
  frequency on logarithmic axes. (d) Simulated Bragg line counts for Cu
  K$\alpha$ radiation over $2\theta\in[5^\circ,90^\circ]$; multiple
  reflections at the same position can be merged into one line. Vertical
  lines mark 128- and 512-peak budgets. The displayed histogram ends at 800
  lines, while the annotated maximum is 7,107.}
  \label{fig:dataset-structural-distributions}
  \end{minipage}
\end{figure}

The main text summarizes source provenance, crystal-system frequencies,
element occurrence, and element-pair frequencies
(\cref{sec:benchmark,fig:dataset-coverage}). Here, we complement that
overview with distributions of primitive-cell size, composition arity
(the number of distinct elements), prototype frequency, and simulated
Bragg line counts
(\cref{fig:dataset-structural-distributions}). All statistics reported
below are computed over the final curated benchmark corpus of 4,262,691
structures. Together, these statistics characterize cell size, chemical
diversity, structural-family coverage, and diffraction-profile complexity.

Primitive cells have a median of 6 atoms, with the 95th percentile at
22 atoms (\cref{fig:dataset-atom-count}). Ternary compositions dominate
the corpus at 69.5\%, followed by quaternary compositions at 20.4\% and
binary compositions at 8.0\%. Compositions with five and six element types
account for 1.7\% and 0.3\%, respectively, while unary compositions and
those with seven or eight element types each account for less than 0.1\%
(\cref{fig:dataset-composition-arity}).

The prototype distribution is strongly long-tailed. Among 46,658 structural
prototypes, 41.4\% occur only once
(\cref{fig:dataset-prototype-frequency}). Thus, despite the large size of
the corpus, coverage across structural families remains highly nonuniform.

The simulated profiles also span a wide range of diffraction complexity
(\cref{fig:dataset-peak-complexity}). Within
$2\theta\in[5^\circ,90^\circ]$, the number of simulated Bragg lines follows
a long-tailed distribution that extends well beyond the 128- and 512-line
budgets indicated in the figure. These counts refer to the peak positions
returned by the diffraction calculator, which can merge multiple reflections
at the same position into a single line. They therefore do not count
individual reflections and need not correspond to the number of
experimentally resolved peaks. The broad distribution motivates examining
how much diffraction information is retained by fixed-size peak
representations.

\newpage
\subsection{Construction of the Multiphase Dataset}
\label{appx:multiphase-dataset}

We construct the multiphase dataset independently within the original
MP-20 train, validation, and test splits, so that no constituent structure
crosses the original split boundaries. We first identify binary metallic
chemical systems and collect split-local unary and binary structures belonging
to each system. Using a frozen Materials Project GGA/GGA+U snapshot, we build
the corresponding binary phase diagrams and enumerate pairs of phases with
distinct compositions.

For each phase pair, we consider nine phase fractions. Let $w_1$ and
$w_2$ denote the weight fractions of the two constituent phases, with
\begin{equation}
    w_1 + w_2 = 1,
    \qquad
    w_1 \in \{0.1,0.2,\ldots,0.9\},
    \qquad
    w_2 = 1-w_1.
\end{equation}
For non-tie mixtures, we define the phase with the larger weight fraction as
the dominant phase and the other as the minor phase. Mixtures with
$w_1=w_2=0.5$ have no dominant--minor assignment.
Each combination of a phase pair and weight fraction defines a candidate
mixture. We retain candidate mixtures that either correspond to equilibrium
two-phase assemblages between adjacent stable phases on the binary convex hull,
or whose mixture energy lies within 50 meV/atom of the convex hull at the
corresponding overall composition. In the latter case, each constituent phase
is also required to satisfy the corresponding stability criterion.

We simulate the constituent PXRD profiles over
$2\theta\in[5^\circ,90^\circ]$ using Cu K$\alpha$ radiation. To convert the
phase weight fractions into diffraction scale factors, let $M_i$ and $V_i$
denote the unit-cell mass and volume of phase $i\in\{1,2\}$. Using the
Hill--Howard relation between phase weight fraction and Rietveld scale factor
~\citep{hill1987quantitative}, we define
\begin{equation}
    \tilde{s}_i = \frac{w_i}{M_i V_i},
    \qquad
    s_i =
    \frac{\tilde{s}_i}{\tilde{s}_1+\tilde{s}_2}.
\end{equation}

Given constituent profiles $x_1(2\theta)$ and $x_2(2\theta)$, we construct
the mixed profile by linear superposition,
\begin{equation}
    \tilde{x}_{\mathrm{mix}}(2\theta)
    =
    s_1 x_1(2\theta) + s_2 x_2(2\theta),
\end{equation}
followed by maximum-intensity normalization,
\begin{equation}
    x_{\mathrm{mix}}(2\theta)
    =
    \frac{\tilde{x}_{\mathrm{mix}}(2\theta)}
         {\max_{2\theta'}\tilde{x}_{\mathrm{mix}}(2\theta')}.
\end{equation}

Each record stores the mixed and constituent profiles, both crystal structures, phase weight fractions, Rietveld scale factors, compositions, and primitive-cell atom counts. The final dataset contains 100,798 training, 12,418 validation, and 11,848 test mixtures (125,064 total). For the reported recovery experiment, we use the 10,657 non-tie test mixtures.
\newpage

\subsection{Evaluation Metrics}
\label{appx:evaluation-metrics}

\paragraph{Top-$K$ match rate.}
We evaluate structure recovery using the Top-$K$ match rate. For a dataset $\mathcal{D}$ containing $|\mathcal{D}|$ target structures, let $C_n$ denote the reference structure for example $n$, and let $\hat{C}_n^{(j)}$ denote the $j$-th generated candidate. We define the per-example Top-$K$ success indicator as
\begin{equation}
m_n^{(K)}
=
\mathbb{I}
\left[
\exists j \in \{1,\ldots,K\}
:
\hat{C}_n^{(j)} \simeq C_n
\right],
\end{equation}
where $\simeq$ denotes structural equivalence under pymatgen's \texttt{StructureMatcher}.
The Top-$K$ match rate is then
\begin{equation}
\mathrm{Top}\text{-}K
=
\frac{1}{|\mathcal{D}|}
\sum_{n=1}^{|\mathcal{D}|}
m_n^{(K)}.
\end{equation}
Thus, an example is counted as successfully recovered if at least one of its first $K$ generated candidates matches the reference structure. For all single-phase evaluations, we use \texttt{StructureMatcher} with $(\texttt{ltol}, \texttt{stol}, \texttt{angle\_tol})=(0.2, 0.3, 5^\circ)$. We report Top-1 and Top-20 match rates throughout the benchmark.

\paragraph{Multiphase recovery metrics.}
For the two-phase benchmark, each example $n$ contains two reference structures, $C_{n,1}$ and $C_{n,2}$. Let $\hat{C}_{n,i}^{(j)}$ denote the $j$-th generated candidate for phase $i \in \{1,2\}$. We define the per-phase Top-$K$ success indicator as
\begin{equation}
m_{n,i}^{(K)}
=
\mathbb{I}
\left[
\exists j \in \{1,\ldots,K\}
:
\hat{C}_{n,i}^{(j)} \simeq C_{n,i}
\right].
\end{equation}

Let $d_n$ and $r_n$ denote the dominant and minor phases of example $n$, respectively. The dominant-phase and minor-phase Top-$K$ match rates are
\begin{equation}
\mathrm{Top}\text{-}K_{\mathrm{dom}}
=
\frac{1}{|\mathcal{D}_{\mathrm{multi}}|}
\sum_{n=1}^{|\mathcal{D}_{\mathrm{multi}}|}
m_{n,d_n}^{(K)},
\end{equation}
and
\begin{equation}
\mathrm{Top}\text{-}K_{\mathrm{min}}
=
\frac{1}{|\mathcal{D}_{\mathrm{multi}}|}
\sum_{n=1}^{|\mathcal{D}_{\mathrm{multi}}|}
m_{n,r_n}^{(K)}.
\end{equation}

Finally, the all-phase Top-$K$ match rate requires both constituent structures to be recovered within their respective candidate sets:
\begin{equation}
\mathrm{Top}\text{-}K_{\mathrm{all}}
=
\frac{1}{|\mathcal{D}_{\mathrm{multi}}|}
\sum_{n=1}^{|\mathcal{D}_{\mathrm{multi}}|}
m_{n,d_n}^{(K)}
m_{n,r_n}^{(K)}.
\end{equation}
The all-phase metric is therefore a joint, sample-level success rate rather than an average of the dominant- and minor-phase match rates. We report dominant-, minor-, and all-phase Top-1 and Top-20 match rates.

\newpage
\subsection{Existing datasets used in additional experiments}

This subsection documents the existing datasets used in our additional
experiments. We describe MP-20, which serves as a compact reference training
corpus for comparisons with prior PXRD-to-structure models, and
MP-20-polymorph-split, which we use for supplementary evaluation under a
composition-disjoint split.

\textbf{MP-20.}
MP-20 is a widely used benchmark for crystal structure prediction derived
from the Materials Project~\citep{jain2013materialsproject,jiao2023diffcsp}.
It contains inorganic crystal structures with at most 20 atoms in the
primitive cell and provides standard train, validation, and test splits.
We use MP-20 as a compact reference training corpus for comparisons with
prior PXRD-to-structure models and for the representation and augmentation
studies reported in the main text and appendix. For experimental PXRD
evaluation, we remove from the training set all structures that structurally
match reference structures in RRUFF or opXRD.

\textbf{MP-20-polymorph-split.}
MP-20-polymorph-split is a composition-grouped repartition of MP-20
introduced by \citet{martirossyan2025structurematches}. Whereas the
standard MP-20 split can place structures with the same reduced composition
in different subsets, MP-20-polymorph-split assigns all structures sharing
a reduced composition to the same split while preserving the overall
composition-arity distribution. It therefore removes composition overlap
between training and evaluation while retaining the underlying MP-20
structure collection.

\newpage
\section{Additional experimental results}
\label{appx:additional-experiments}

\subsection{PXRD representation study}
\label{appx:pxrd-representation-study}

We study the design choices for PXRD input representation and encoding in
three stages. First, we compare alternative representations of the physical
peak position using $2\theta$, $d$-spacing, and $q$-spacing, with intensity as
the only additional peak feature. Second, starting from the best peak-position
representation, we examine the effect of adding peak width and prominence.
Third, we augment the resulting peak features with Fourier encodings of the
physical peak position and compare different frequency ranges. Throughout the
study, we keep the model architecture and size fixed, train each model on
MP-20 for 200K steps, and use our full PXRD profile augmentation. We compare
the representations using Top-1 match rates on the RRUFF and opXRD validation
sets. Results are summarized in \Cref{tab:pxrd-representation-ablation}.












\begin{table}[htbp]
\caption{Top-1 validation match rates (\%) for ablations of the PXRD input
representation and peak-position encoding. The first three rows compare
peak-position representations using intensity only. Subsequent rows
cumulatively augment the $q,I$ representation with peak prominence and width,
followed by Fourier embeddings of the peak position.}
\label{tab:pxrd-representation-ablation}
\centering
\setlength{\tabcolsep}{8pt}
\begin{tabular}{lccc}
\toprule
\textbf{Input feature} & \textbf{MP-20} & \textbf{RRUFF} & \textbf{opXRD} \\
\midrule

\multicolumn{4}{l}{
\cellcolor{gray!8}\textit{Peak-position representation}} \\

$2\theta, I$
& 60.94
& 36.67\std{1.86}
& 36.67\std{6.18} \\

$d, I$
& 59.96
& 39.17\std{2.28}
& 35.83\std{2.28} \\

$q, I$
& 64.84
& 41.25\std{1.74}
& 37.50\std{2.95} \\

\cmidrule(lr){1-4}
\multicolumn{4}{l}{
\cellcolor{gray!8}\textit{Additional peak features}} \\

$+$ prominence, width
& 66.21
& 43.33\std{2.28}
& 41.67\std{4.17} \\

\cmidrule(lr){1-4}
\multicolumn{4}{l}{
\cellcolor{gray!8}\textit{Peak-position Fourier embedding}} \\

$+$ log16
& \textbf{69.99}
& \textbf{47.92}\std{1.25}
& \textbf{45.56}\std{3.94} \\

$+$ log64
& 69.73
& 44.86\std{1.46}
& 40.83\std{2.50} \\

\bottomrule
\end{tabular}
\end{table}

\textbf{Peak-position representation.}
We first compare $2\theta$, $d$-spacing, and $q$-spacing as representations of
the physical peak position, using peak intensity $I$ as the only additional
feature. Among the three representations, $(q,I)$ gives the strongest
validation performance overall. Unlike $2\theta$, both $d$- and $q$-spacing
are independent of the measurement wavelength, avoiding the need to represent
the same scattering geometry differently across measurement wavelengths.
Based on the validation results, we use $q$ as the peak-position
representation for the remainder of the study.

\textbf{Peak-shape features.}
Starting from the $(q,I)$ representation, we additionally provide the width
$w$ and prominence $p$ of each detected peak. These descriptors characterize
local peak shape and distinguish peaks that may have similar positions and
intensities but different profiles. Adding width and prominence improves
validation performance on both RRUFF and opXRD, indicating that peak-shape
information provides useful conditioning beyond peak positions and
intensities alone. We therefore use $(q,I,w,p)$ as the base peak
representation for the following experiments.

\textbf{Fourier encoding of physical peak positions.}
We further augment the representation with Fourier features of the physical peak position $q$. This encoding operates on the value of $q$ itself rather than on the index of a peak in the input sequence, and is therefore distinct from positional encoding for token order. We use 16 logarithmically spaced frequencies,
\begin{equation}
    f_k = F_{\max}^{k/15},
    \qquad k = 0,\ldots,15,
\end{equation}
where $F_{\max}=16$ and $F_{\max}=64$ define the \textit{log16} and \textit{log64} settings, respectively. Thus, both settings use 16 frequencies spanning $1$ to $F_{\max}$ on a logarithmic scale. Further details of the Fourier encoding and peak embedding are provided in \cref{appx:pxrd-encoding}.

As shown in \Cref{tab:pxrd-representation-ablation}, adding Fourier features
with the log16 setting improves validation performance on both RRUFF and
opXRD and gives the best overall results. Increasing the maximum frequency to
64 reduces performance on both datasets. We therefore use the log16 Fourier
encoding in the final model.

\subsection{Architecture ablation}
\label{appx:architecture-ablation}
We ablate three architectural components of \model{}: in-context
conditioning, the PXRD encoder architecture, and the Geometry Enhancement
Module (GEM). To isolate architectural effects from simulation-to-real
robustness, we train all models on MP-20 for 300K steps without PXRD
augmentation and evaluate them on the clean simulated MP-20 validation set.

\begin{table}[htbp]
\caption{Architecture component study on the MP-20 validation set.
Top-$K$ match rates (\%) are reported as mean $\pm$ sample standard deviation
over three training seeds. All models are evaluated at the final 300K-step
EMA checkpoint.}
\label{tab:architecture-ablation}
\centering
\begingroup
\setlength{\tabcolsep}{8pt}
\begin{tabular}{lccc}
\toprule
& \textbf{Top-1} & \textbf{Top-20}
& \boldmath$\Delta$\unboldmath\ \textbf{Top-1 (pp)} \\
\midrule

\model{}
& 78.86\std{0.40}
& 90.62\std{0.40}
& -- \\

$-$ In-context fusion
& 78.64\std{0.34}
& 90.33\std{0.28}
& $-0.22$ \\

$-$ GEM
& 76.35\std{0.59}
& 89.90\std{0.05}
& $-2.51$ \\

\bottomrule
\end{tabular}
\endgroup
\end{table}

\begin{table}[htbp]
\caption{PXRD encoder study on the MP-20 validation set.
Values are Top-$K$ match rates (\%).}
\label{tab:pxrd-encoder-ablation}
\centering
\begingroup
\setlength{\tabcolsep}{8pt}
\begin{tabular}{lcc}
\toprule
\textbf{PXRD encoder} & \textbf{Top-1} & \textbf{Top-20} \\
\midrule

Transformer encoder
& 75.37
& 89.04 \\

MLP encoder
& 72.70
& 87.96 \\

\bottomrule
\end{tabular}
\endgroup
\end{table}

\textbf{In-context fusion and GEM.}
\Cref{tab:architecture-ablation} evaluates the contribution of in-context
PXRD conditioning and GEM. Removing in-context fusion produces a small
reduction in both Top-1 and Top-20 match rates, whereas removing GEM leads
to a larger drop, particularly in Top-1 performance. These
results indicate that the geometry-aware attention bias provided by GEM is
important for structure recovery, while in-context fusion provides a
smaller additional improvement.

\textbf{PXRD encoder architecture.}
\Cref{tab:pxrd-encoder-ablation} compares Transformer and MLP encoders for
the PXRD peak sequence. The Transformer encoder improves both Top-1 and
Top-20 match rates over the MLP encoder, supporting the use of a
sequence-modeling architecture for constructing the PXRD representation
used to condition the structure generator.

\subsection{In-depth analysis of benchmark results}
\begin{figure*}[t]
    \centering
    \includegraphics[width=\textwidth]{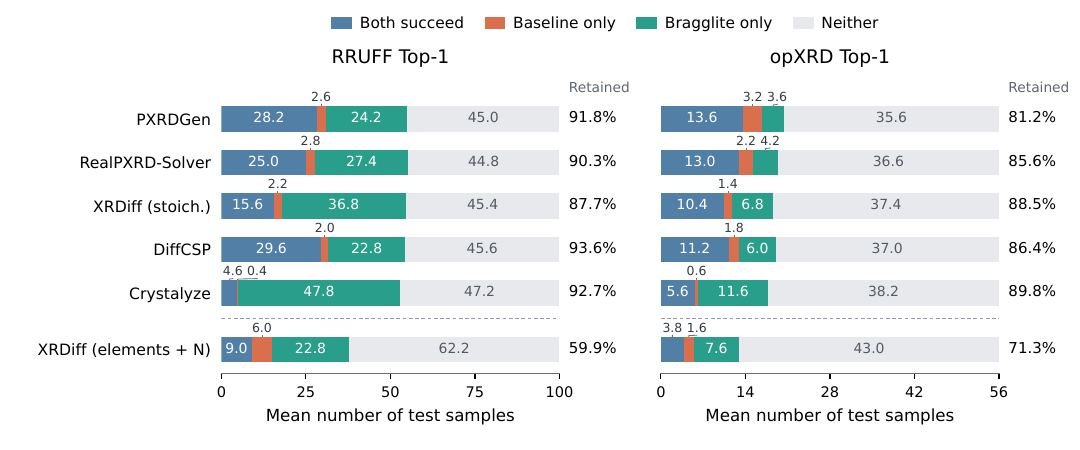}
    \caption{
        Overlap of Top-1 successes between \model{} and prior
        baselines on the RRUFF and opXRD test sets.
        Each bar partitions the test samples into cases where both models
        succeed, only the baseline succeeds, only \model{} succeeds,
        or neither succeeds.
        \emph{Retained} denotes the fraction of baseline successes that are
        also recovered by \model{}.
    }
    \label{fig:baseline-success-overlap}
\end{figure*}
\textbf{Overlap of successful predictions.}
To better understand whether the improvement of \model{} comes from solving
different test cases or consistently recovering those solved by prior methods,
we analyze the overlap of Top-1 successes on the RRUFF and opXRD test sets
in \Cref{fig:baseline-success-overlap}.
For each baseline, we partition test samples according to whether both models
succeed, only the baseline succeeds, only \model{} succeeds, or neither
succeeds.
Across most baselines, \model{} recovers the large majority of samples solved
by the baseline while additionally solving a substantially larger set of
previously unsuccessful cases.
This trend is consistent across both experimental datasets, suggesting that
the improved aggregate match rate primarily reflects additional structure
recoveries rather than a trade-off in which previously solved samples are lost.
The overlap is lower for XRDiff under element-set conditioning, indicating
greater complementarity between the two models in this more weakly
conditioned setting.

\subsection{Detailed two-phase recovery results}
\label{sec:two-phase-detailed}
We further report phase-wise recovery results for the two-phase benchmark, separating dominant-phase, minor-phase, and joint all-phase recovery. \Cref{tab:multiphase-detailed} complements the all-phase results in \cref{sec:multiphase-results} by showing how often each constituent phase is recovered individually.

\begin{table}[h]
\caption{
\textbf{Detailed two-phase recovery results.}
Dominant-phase, minor-phase, and all-phase Top-$K$ match rates (\%) on the
MP-20 and GALAXI two-phase benchmarks. All-phase recovery requires both
constituent phases to be recovered within their respective candidate sets.
}
\label{tab:multiphase-detailed}
\centering
\setlength{\tabcolsep}{5pt}
\begin{tabular}{llcccccc}
\toprule
& &
\multicolumn{2}{c}{\textbf{Dominant}}
& \multicolumn{2}{c}{\textbf{Minor}}
& \multicolumn{2}{c}{\textbf{All-phase}} \\
\cmidrule(lr){3-4}
\cmidrule(lr){5-6}
\cmidrule(lr){7-8}
\textbf{Dataset}
& \textbf{Model}
& Top-1 & Top-20
& Top-1 & Top-20
& Top-1 & Top-20 \\
\midrule

\multirow{3}{*}{MP-20}
& MatterGen
& 14.28 & \textbf{55.69}
& 15.48 & \textbf{55.41}
& 2.13 & \textbf{30.67} \\
& XRDiff
& 25.28 & 38.28
& 23.14 & 36.48
& 6.76 & 14.64 \\
& \model{}
& \textbf{33.99} & 48.78
& \textbf{30.59} & 45.02
& \textbf{13.26} & 25.34 \\
\midrule

\multirow{3}{*}{GALAXI}
& MatterGen
& 0.00 & \textbf{76.19}
& 4.76 & \textbf{71.43}
& 0.00 & 42.86 \\
& XRDiff
& 28.57 & 38.10
& 9.52 & 14.29
& 0.00 & 9.52 \\
& \model{}
& \textbf{57.14} & \textbf{76.19}
& \textbf{33.33} & 66.67
& \textbf{19.05} & \textbf{47.62} \\
\bottomrule
\end{tabular}
\vspace{-.1in}
\end{table}

\subsection{Benchmark results on MP-20-polymorph-split}
\label{appx:mp20-polymorph-split-results}

\begin{table}[h]
\caption{Top-$K$ match rates (\%) for models trained on
MP-20-polymorph-split, following the MP-20 configuration in
\Cref{tab:benchmark}.}
\label{tab:benchmark-mp20-polymorph-split}
\centering
\fontsize{7.5}{9}\selectfont
\setlength{\tabcolsep}{2pt}
\renewcommand{\arraystretch}{1.16}
\begin{tabular*}{\textwidth}{@{\hspace{4pt}\extracolsep{\fill}}clcccccc@{\hspace{4pt}}}
\toprule
& & \multicolumn{2}{c}{\textbf{Simulated PXRD}}
& \multicolumn{2}{c}{\textbf{RRUFF}}
& \multicolumn{2}{c}{\textbf{opXRD}} \\
\cmidrule(lr){3-4}\cmidrule(lr){5-6}\cmidrule(lr){7-8}
\textbf{Dataset} & \textbf{Model} & Top-1 & Top-20 & Top-1 & Top-20 & Top-1 & Top-20 \\
\midrule

\multirow{11}{*}{\shortstack{MP-20-\\polymorph-\\split}}
& \multicolumn{7}{l}{\cellcolor{gray!8}\textit{Chemical input: exact stoichiometry}} \\

& PXRDGen
& 55.43 & 78.92
& 30.00\std{3.08} & 59.00\std{1.41}
& 24.29\std{1.60} & 36.79\std{2.04} \\

& RealPXRD-Solver
& 57.42 & 79.54
& 26.60\std{2.61} & 47.80\std{2.68}
& 24.64\std{1.49} & 33.21\std{1.60} \\

& XRDiff
& 60.26 & 77.54
& 23.00\std{2.35} & 45.60\std{1.52}
& 18.93\std{2.71} & 29.29\std{1.60} \\

& Crystalyze
& 31.07 & 49.46
& \phantom{0}2.00\std{1.41} & 11.00\std{1.22}
& 11.79\std{0.98} & 17.50\std{0.80} \\

& DiffCSP (no PXRD)
& 51.85 & 75.71
& 29.00\std{3.16} & 58.60\std{1.52}
& 20.71\std{1.60} & 36.07\std{0.80} \\

& Crystalite (no PXRD)
& 62.28 & 79.45
& 41.20\std{3.56} & 60.40\std{1.82}
& 25.36\std{1.49} & 35.71\std{1.26} \\

& \textbf{\model{}}
& \textbf{68.43} & \textbf{81.48}
& \textbf{49.00}\std{1.22} & \textbf{65.40}\std{1.34}
& \textbf{27.86}\std{2.04} & \textbf{38.57}\std{2.04} \\

\cmidrule(lr){2-8}
& \multicolumn{7}{l}{\cellcolor{gray!8}\textit{Chemical input: element set + $N$}} \\

& XRDiff
& 35.92 & 50.79
& 15.60\std{1.52} & 26.60\std{1.67}
& 10.36\std{1.49} & 17.14\std{0.98} \\

& \textbf{\model{}}
& \textbf{44.20} & \textbf{69.76}
& \textbf{30.60}\std{1.52} & \textbf{55.20}\std{1.92}
& \textbf{17.86}\std{1.26} & \textbf{23.93}\std{1.60} \\

\bottomrule
\end{tabular*}
\end{table}

\Cref{tab:benchmark-mp20-polymorph-split} follows the model lineup,
chemical-input settings, and evaluation columns of the MP-20 block in
\Cref{tab:benchmark}, using MP-20-polymorph-split as the training dataset
and its held-out test split for simulated PXRD evaluation.

\clearpage
\section{\model{} model details}
\label{appx:model-details}

This section provides additional implementation details for \model{}.
We first summarize the Crystalite backbone, then describe the PXRD
representation, encoder, in-context conditioning, and augmentation pipeline.
Finally, we describe how \model{} is combined with XDecomposer for two-phase
structure recovery.

\model{} builds on Crystalite~\citep{veljkovic2026crystalite} and preserves
its diffusion state, EDM formulation, sampler, and structural prediction heads.
To condition the denoiser on diffraction, we introduce a PXRD encoder that maps
the observed peak list to a short sequence of context tokens. These tokens are
concatenated with the structural tokens and jointly processed by the Crystalite
Transformer, allowing the observed diffraction pattern to influence structural
predictions throughout denoising.

\subsection{Crystalite backbone}

Crystalite~\citep{veljkovic2026crystalite} is a geometry-aware diffusion
Transformer for crystal generation. It represents a unit cell as a continuous
state consisting of atom-type features, fractional coordinates, and a
six-dimensional lattice latent. Each atomic site is represented by a single
token that combines its atom-type and fractional-coordinate embeddings, while
the lattice is represented by a separate global token. The resulting sequence
is processed by a noise-conditioned Transformer under the EDM denoising
formulation.

To explicitly incorporate periodic crystal geometry, Crystalite uses the
Geometry Enhancement Module (GEM). GEM computes minimum-image pair geometry
from the current fractional coordinates and lattice and injects the resulting
features into atom--atom attention as additive geometric biases. This provides
the Transformer with explicit lattice-dependent periodic geometry during
denoising.

For atom types, Crystalite uses Subatomic Tokenization, which represents each
element using chemically structured descriptors derived from its periodic-table
position and electronic configuration. These descriptors are projected onto a PCA basis and normalized to form
continuous atom-type features. During de novo
generation, atom-type features, coordinates, and lattice variables are jointly
denoised, and the predicted atom-type features are mapped back to discrete
elements by nearest-prototype decoding. In the CSP setting, atom types are fixed
and only the coordinates and lattice are denoised.

These two operating modes align naturally with our benchmark settings. Under
exact-stoichiometry conditioning, atom types are fixed throughout generation.
Under element-set conditioning, atom-type features are additionally denoised
and decoded while restricting the output elements to the supplied element set.

\subsection{PXRD encoding and in-context conditioning}
\label{appx:pxrd-encoding}

\textbf{Peak extraction.}
For both simulated and experimental PXRD, we begin with a dense intensity
profile
\begin{equation}
  \mathbf{x}
  =
  (x_1,\ldots,x_M)
  \in \mathbb{R}_{\geq 0}^{M},
  \qquad M=850,
\end{equation}
sampled at $0.1^\circ$ intervals over
$2\theta\in[5^\circ,90^\circ)$.
We detect peaks from $\mathbf{x}$ using
\texttt{scipy.signal.find\_peaks}~\citep{virtanen2020scipy} with
\texttt{prominence=0.0} and no additional minimum-height or
minimum-distance constraint. Peak widths are computed using \texttt{scipy.signal.peak\_widths}
at relative height $0.5$ and converted from bins to the units of the
corresponding peak-position representation.

If more than 128 peaks are detected, we retain the 128 peaks with the highest
intensities and sort them by increasing $2\theta$. Let $K\leq128$ denote the
number of retained peaks. We represent the resulting peak sequence as
\begin{equation}
  \mathcal{P}_x
  =
  (\mathbf{p}_1,\ldots,\mathbf{p}_K),
  \qquad
  \mathbf{p}_k
  =
  \left(q_k,I_k,w_k,\pi_k\right)
  \in\mathbb{R}^{4},
  \label{eq:model-peak-sequence}
\end{equation}
where $q_k$, $I_k$, $w_k$, and $\pi_k$ denote the scattering-vector magnitude,
normalized intensity, peak width, and prominence, respectively. Let
$\widetilde{I}_k$ denote the unnormalized intensity of peak $k$. We normalize peak intensities by the maximum value of the full profile,
\begin{equation}
  I_k
  =
  \frac{\widetilde{I}_k}{\widetilde{I}_{\max}},
  \qquad
  \widetilde{I}_{\max}
  =
  \max_j x_j.
  \label{eq:model-intensity-normalization}
\end{equation}

\textbf{Fourier encoding for peak positions.}
To provide a richer representation of the physical peak position, we
additionally encode $q_k$ using sinusoidal Fourier features. We first convert
$q_k$ to the dimensionless coordinate
\begin{equation}
  \xi_k
  =
  \frac{q_k\lambda_{\mathrm{ref}}}{4\pi},
  \qquad
  \lambda_{\mathrm{ref}}=1.54184~\text{\AA}.
  \label{eq:model-q-normalization}
\end{equation}
We then use 16 logarithmically spaced frequencies
\begin{equation}
  f_m
  =
  F_{\max}^{m/15},
  \qquad
  m=0,\ldots,15.
  \label{eq:model-fourier-frequencies}
\end{equation}
For each peak, the corresponding
32-dimensional Fourier feature is
\begin{equation}
  \boldsymbol{\phi}(q_k)
  =
  \left[
    \sin(2\pi f_m\xi_k),
    \cos(2\pi f_m\xi_k)
  \right]_{m=0}^{15}
  \in\mathbb{R}^{32}.
  \label{eq:model-fourier-features}
\end{equation}
This encoding provides multiple frequency scales for representing variations
in physical peak position while retaining $q_k$ itself in the base peak
descriptor.

\textbf{Peak encoding.}
The four base peak features are jointly embedded with a two-layer MLP, while
the normalized Fourier feature is projected through a separate linear layer.
The two branches are summed to obtain the peak embedding,
\begin{equation}
  \mathbf{h}_k
  =
  \operatorname{MLP}_{\mathrm{peak}}(\mathbf{p}_k)
  +
  \mathbf{W}_{F}
  \operatorname{Normalize}
  \left(
    \boldsymbol{\phi}(q_k)
  \right)
  \in\mathbb{R}^{d_x},
  \qquad
  d_x=256,
  \label{eq:model-peak-embedding}
\end{equation}
where
\begin{equation}
  \operatorname{MLP}_{\mathrm{peak}}
  =
  \operatorname{Linear}_{d_x\rightarrow d_x}
  \circ
  \operatorname{SiLU}
  \circ
  \operatorname{Linear}_{4\rightarrow d_x},
\end{equation}
and $\mathbf{W}_{F}\in\mathbb{R}^{d_x\times32}$ denotes the bias-free Fourier
projection.

\textbf{Profile encoding.}
We prepend a learned \texttt{[CLS]} embedding
$\mathbf{c}_{\mathrm{CLS}}\in\mathbb{R}^{d_x}$ to the peak embeddings and
process the resulting sequence using a four-layer, eight-head Transformer:
\begin{equation}
  \mathbf{U}_x
  =
  \operatorname{PeakTransformer}
  \left(
    [\mathbf{c}_{\mathrm{CLS}};
     \mathbf{h}_1;
     \ldots;
     \mathbf{h}_K]
  \right)
  \in\mathbb{R}^{(K+1)\times d_x}.
  \label{eq:model-peak-transformer}
\end{equation}
The output corresponding to the \texttt{[CLS]} token provides a global
representation of the observed PXRD profile. We project this representation
to the hidden dimension of the Crystalite backbone,
\begin{equation}
  \mathbf{z}_x
  =
  \operatorname{Linear}_{\mathrm{proj}}
  \left(
    \mathbf{U}_{x,\mathrm{CLS}}
  \right)
  \in\mathbb{R}^{d}.
  \label{eq:model-pxrd-summary}
\end{equation}

\textbf{In-context conditioning.}
We form $R=8$ context tokens by repeating the global PXRD representation and
adding a distinct learned positional embedding to each copy:
\begin{equation}
  \mathbf{Z}_x
  =
  \left[
    \mathbf{z}_x+\mathbf{e}_1;
    \ldots;
    \mathbf{z}_x+\mathbf{e}_R
  \right]
  \in\mathbb{R}^{R\times d},
  \qquad R=8,
  \label{eq:model-context}
\end{equation}
where $\mathbf{e}_r\in\mathbb{R}^{d}$ is a learned positional embedding for
the $r$-th context token. We prepend these context tokens to the original Crystalite structural token
sequence,
\begin{equation}
  \mathbf{T}^{(0)}
  =
  [\mathbf{Z}_x;\mathbf{T}_{\mathrm{str}}^{(0)}]
  \in\mathbb{R}^{(R+N+1)\times d},
  \label{eq:model-conditioned-sequence}
\end{equation}
where $\mathbf{T}_{\mathrm{str}}^{(0)}\in\mathbb{R}^{(N+1)\times d}$ is the original Crystalite structural sequence, consisting of the $N$ atomic-site tokens and one global lattice token. The PXRD context depends only on the observed profile and is therefore reused
across diffusion steps, whereas $\mathbf{T}_{\mathrm{str}}^{(0)}$ is
recomputed from the current noisy crystal. Within each denoiser evaluation,
the context and structural tokens are jointly processed by the Crystalite
Transformer, allowing the atom and lattice tokens to attend to the diffraction
representation throughout denoising.

\subsection{PXRD augmentation}
\label{appx:pxrd-augmentation}

\textbf{Overview.}
For simulated training data, we augment PXRD profiles online to mimic
distortions commonly observed in experimental measurements. Starting from Bragg reflections computed for each crystal structure under Cu K$\alpha$ radiation ($\lambda=1.54184$~\AA), we first apply reflection-level
perturbations to peak presence, position, and intensity. We then render the
perturbed reflections into a dense profile using randomized pseudo-Voigt peak
shapes and apply profile-level background and noise augmentation. The resulting
profile is converted to the peak representation using the extraction procedure
described in \cref{appx:pxrd-encoding}. All random augmentation parameters are
resampled whenever a training example is loaded. 

\textbf{Weak-peak removal.}
Weak-reflection removal has been used in prior PXRD augmentation schemes,
including RealPXRD-Solver~\citep{li2026realpxrdsolver}, to model weak peaks
that may be missing or unresolved in experimental patterns. We randomly remove a fraction of
the lowest-intensity Bragg reflections during training. Let the simulated
pattern contain $M$ reflections. We sample
\begin{equation}
    r_{\mathrm{drop}}
    \sim
    \mathcal{U}(0,r_{\mathrm{drop}}^{\max}),
    \qquad
    n_{\mathrm{drop}}
    =
    \left\lfloor r_{\mathrm{drop}}M \right\rfloor,
\end{equation}
and remove the $n_{\mathrm{drop}}$ weakest reflections while retaining at
least one reflection. In our experiments, we use
$r_{\mathrm{drop}}^{\max}=0.2$.

\textbf{Peak-position jitter.} 
Following prior experiment-inspired PXRD augmentation
~\citep{li2026realpxrdsolver}, we independently perturb the position of each
retained reflection to model local peak-position errors:
\begin{equation}
    2\theta_i'
    =
    2\theta_i+\delta_i,
    \qquad
    \delta_i
    \sim
    \mathcal{U}(-\Delta_{\mathrm{jitter}},
                 \Delta_{\mathrm{jitter}}).
\end{equation}
Unlike the shared zero shift described below, these perturbations are sampled
independently for individual reflections. We use
$\Delta_{\mathrm{jitter}}=0.2^\circ$.

\textbf{Peak-intensity scaling.}
Prior PXRD augmentation schemes perturb relative peak intensities to improve
robustness to experimental intensity distortions, including effects associated
with preferred orientation~\citep{li2026realpxrdsolver}. We independently
scale each retained reflection as
\begin{equation}
    I_i'
    =
    s_i I_i,
    \qquad
    s_i
    \sim
    \mathcal{U}(s_{\min},s_{\max}).
\end{equation}
The perturbed intensities are then jointly rescaled so that their maximum
matches that of the original reflection list. We use
$(s_{\min},s_{\max})=(0.6,1.4)$.

This independent intensity scaling provides a simple approximation to
experimental variations in relative peak intensities, including those that
may arise from preferred orientation, but does not constitute a
crystallographic preferred-orientation model because it does not depend on
reflection orientation.

\textbf{Zero shift.}
Global angular shifts have been used in physics-informed XRD augmentation to
model systematic measurement offsets~\citep{oviedo2019fast}. We apply a
single shift shared by all reflections in a profile,
\begin{equation}
    z
    \sim
    \mathcal{U}(-\Delta_{\mathrm{zero}},
                 \Delta_{\mathrm{zero}}),
    \qquad
    2\theta_i''
    =
    2\theta_i'+z.
\end{equation}
This perturbation models coherent offsets such as zero-point calibration or
sample-displacement errors, in contrast to the independent peak-position
jitter above. We use
$\Delta_{\mathrm{zero}}=0.3^\circ$.

\textbf{Pseudo-Voigt broadening.}
Prior PXRD models have incorporated peak broadening and profile-level
augmentation to improve robustness to experimental diffraction
patterns~\citep{segal2026xrdiff}. We render the perturbed
Bragg reflections into a dense profile using pseudo-Voigt peak shapes.

The base full width at half maximum of reflection $i$ follows the Caglioti form
\begin{equation}
    H_i^{(0)}
    =
    \sqrt{
        U\tan^2\theta_i
        +V\tan\theta_i
        +W
    }.
\end{equation}
We use
\begin{equation}
    (U,V,W)=(0.1,0.01,0.1).
\end{equation}
To vary the overall broadening across profiles, we sample a profile-wide width
scale
\begin{equation}
    s_w
    \sim
    \mathcal{U}(s_w^{\min},s_w^{\max}),
    \qquad
    H_i=s_w H_i^{(0)}.
\end{equation}
We use
$(s_w^{\min},s_w^{\max})=(0.4,2.2)$.

We additionally sample one pseudo-Voigt mixing coefficient per profile,
\begin{equation}
    \eta
    \sim
    \mathcal{U}(0,\eta_{\max}),
\end{equation}
with $\eta_{\max}=0.8$, and render each reflection as
\begin{equation}
\begin{split}
    P_i(t)
    =
    I_i'
    \Bigg[
    &(1-\eta)
    \exp\left(
        -4\ln 2
        \left(
            \frac{t-2\theta_i''}{H_i}
        \right)^2
    \right) \\
    &+
    \eta
    \frac{1}{
        1+4
        \left(
            \frac{t-2\theta_i''}{H_i}
        \right)^2
    }
    \Bigg],
\end{split}
\end{equation}
where $t$ denotes the $2\theta$ coordinate. The broadened profile is
$P(t)=\sum_i P_i(t)$.

\textbf{Background augmentation.}
Smooth background variation is a common experimental artifact and has been
included in physics-informed XRD augmentation pipelines
~\citep{oviedo2019fast}. We normalize the profile coordinate to
$u\in[-1,1]$ and add a quadratic background
\begin{equation}
    B(u)
    =
    A\max\left(0,b_0+b_1u+b_2u^2\right).
\end{equation}
We sample
\begin{equation}
\begin{aligned}
    A   &\sim \mathcal U(0,A_{\max}), &
    b_0 &\sim \mathcal U(0,1), \\
    b_1 &\sim \mathcal U(-0.5,0.5), &
    b_2 &\sim \mathcal U(0,0.5),
\end{aligned}
\end{equation}
and use $A_{\max}=0.35$.

\textbf{Gaussian noise.}
Additive noise is also commonly used to approximate measurement noise in
experiment-inspired PXRD augmentation
~\citep{oviedo2019fast,li2026realpxrdsolver}. We sample one noise scale per
profile,
\begin{equation}
    \sigma_n
    \sim
    \mathcal U(0,\sigma_n^{\max}),
    \qquad
    \epsilon_j
    \sim
    \mathcal N(0,\sigma_n^2),
\end{equation}
and use $\sigma_n^{\max}=0.04$. The final augmented profile is
\begin{equation}
    \widetilde P_j
    =
    \operatorname{normalize}_{\max}
    \left[
        \max\left(0,P_j+B_j+\epsilon_j\right)
    \right].
\end{equation}

The resulting profile is processed using the same peak-extraction and encoding
pipeline described in \cref{appx:pxrd-encoding}.

\subsection{Multiphase recovery pipeline}
\label{sec:multiphase-extension}

We construct the two-phase recovery pipeline by fine-tuning both
XDecomposer~\citep{gao2026xdecomposer} and the element-set-conditioned
\model{} on the simulated multiphase training set.

\textbf{Ordered phase decomposition.}
Starting from the publicly available XDecomposer checkpoint, we fine-tune the
model to produce an ordered decomposition of each mixed PXRD profile, with the
first output corresponding to the dominant phase and the second to the minor
phase. Given a mixed profile $\mathbf{x}_{\mathrm{mix}}$, the fine-tuned
decomposer predicts
\begin{equation}
    (\mathbf{x}_1,\mathbf{x}_2)
    =
    \operatorname{XDecomposer}(\mathbf{x}_{\mathrm{mix}}),
\end{equation}
where $\mathbf{x}_1$ and $\mathbf{x}_2$ are the predicted dominant- and
minor-phase profiles, respectively. This ordering allows each decomposed
profile to be paired with the oracle primitive-cell atom count of the
corresponding phase, denoted by $N_1$ and $N_2$.

\textbf{Phase-wise structure recovery.}
We additionally fine-tune the element-set-conditioned \model{} on the
multiphase training set so that it predicts each constituent structure from a
phase-specific PXRD profile while receiving only mixture-level chemical
information. Let
\begin{equation}
    \mathcal{E}_{\mathrm{mix}}
    =
    \mathcal{E}_1 \cup \mathcal{E}_2
\end{equation}
denote the element set of the full mixture. For each phase
$i\in\{1,2\}$, the structure generator receives the decomposed profile
$\mathbf{x}_i$, the shared mixture-level element set
$\mathcal{E}_{\mathrm{mix}}$, and the oracle atom count $N_i$:
\begin{equation}
    C_i
    \sim
    p_\theta
    \left(
        \cdot
        \mid
        \mathbf{x}_i,
        \mathcal{E}_{\mathrm{mix}},
        N_i
    \right).
\end{equation}
The two phases are generated independently using the same fine-tuned model.

At evaluation time, we therefore first decompose the mixed profile into
ordered dominant- and minor-phase profiles, pair them with $N_1$ and $N_2$,
respectively, and then apply \model{} independently to recover the two
structures. Importantly, the structure generator is not given phase-specific
element sets or compositions; it must infer which elements from
$\mathcal{E}_{\mathrm{mix}}$ belong to each phase, together with their
multiplicities, atomic coordinates, and lattice.

\textbf{Reference methods.}
For XRDiff, we use the same ordered XDecomposer outputs as for
\model{} and apply the structure generator independently to the
predicted dominant- and minor-phase profiles.
XRDiff additionally receives the phase-specific element set
$\mathcal{E}_i$ and oracle atom count $N_i$ for each phase, since we
do not fine-tune XRDiff to condition on the mixture-level element set.

MatterGen instead receives the shared mixture-level element set
$\mathcal{E}_{\mathrm{mix}}$ and oracle atom count $N_i$ for each
phase, but does not condition on the decomposed PXRD profiles.
For each phase and each reference method, we generate 20 candidate
structures and evaluate dominant-, minor-, and all-phase recovery
using the protocol in \cref{appx:evaluation-metrics}.

\newpage
\section{Experimental details}
\label{appx:experiment-details}

In this section, we provide detailed experimental settings for the results reported in the main text, including our model architecture, training and evaluation configurations, and the training and evaluation settings of the benchmark baselines.

\subsection{Model configurations}
\label{appx:model-configuration}

We train separate \model{} models for the known-composition and element-set settings. All models use Crystalite's PCA-16 subatomic type representation and Geometry Enhancement Module (GEM). PXRD observations are represented by up to 128 intensity-selected peaks, with peak extraction and preprocessing detailed in \cref{appx:pxrd-encoding}.

\textbf{PXRD encoder.}
We use a four-layer Transformer with hidden dimension 256 and eight attention heads as our PXRD encoder. Each peak is represented by $(q, I, \Delta q, \mathrm{prominence})$, where $q=2\pi/d$ and $\Delta q$ is obtained by converting the detected angular peak width to $q$ space using the Jacobian at the peak center. We additionally encode $q$ using 16 logarithmically spaced Fourier frequencies with $F_{\max}=16$, yielding a 32-dimensional Fourier encoding. The encoder uses \texttt{[CLS]} pooling and a padding mask for variable-length peak sequences, with no dropout. The resulting PXRD embedding is projected to the Crystalite backbone width and injected into the token sequence for in-context conditioning.

\textbf{Backbone architecture.}
We use the same \model{} backbone configuration for models trained on PXRD-Bench and MP-20, with hidden dimension 512, 14 Transformer layers, and 16 attention heads. Each Transformer block uses a feed-forward expansion ratio of 4 with GELU activations, pre-LayerNorm without affine parameters, and no attention or feed-forward dropout. The backbone also applies an affine-free LayerNorm before the output heads. Time conditioning is incorporated through adaLN-Zero, while PXRD conditioning is provided in context as described above.

\textbf{Coordinate embedding.}
Fractional atomic coordinates are first wrapped to $[0,1)$ and encoded independently along each coordinate axis using integer Fourier frequencies $k=1,\ldots,32$:
\begin{equation}
    \phi(\mathbf{f})
    =
    \big[
    \sin(2\pi k f_j),
    \cos(2\pi k f_j)
    \big]_{j=1,2,3;\,k=1,\ldots,32}.
\end{equation}
This produces a 192-dimensional feature vector, which is mapped to the backbone hidden dimension by
$192 \rightarrow 512 \rightarrow 512$ with a SiLU activation between the two linear layers. This coordinate encoding is separate from the Fourier encoding used for PXRD peak positions.

\textbf{Noise conditioning.}
Following the EDM parameterization, we first transform the noise level as
\begin{equation}
    c_{\mathrm{noise}} = \frac{1}{4}\log \sigma.
\end{equation}
We encode $c_{\mathrm{noise}}$ using 128 fixed sinusoidal frequencies
$\omega_j=10000^{-j/128}$ for $j=0,\ldots,127$, producing a 256-dimensional embedding from
$\sin(2\pi c_{\mathrm{noise}}\omega_j)$ and
$\cos(2\pi c_{\mathrm{noise}}\omega_j)$.
In our single-$\sigma$ setting, the same 256-dimensional embedding is supplied to both noise-input slots and concatenated into a 512-dimensional vector. A two-layer MLP,
$512 \rightarrow 512 \rightarrow 512$, with a SiLU activation maps this vector to the noise-conditioning embedding used for adaLN-Zero modulation in each Transformer block.

\textbf{Geometry Enhancement Module.}
We use a single GEM shared across all Transformer layers. GEM computes head-specific attention biases once from the current noisy coordinates, lattice, and noise level, and reuses them across all 14 backbone layers. The distance bias, edge-bias MLP, and noise gate are therefore shared across layers.

\textbf{Atom ordering and positional encoding.}
For the element-set model, we follow the atom-ordering scheme introduced in MCFlow~\citep{seong2026multimodal}, which organizes atoms according to their symmetry orbits. We first identify symmetry orbits using \texttt{SpacegroupAnalyzer} with $\texttt{symprec}=0.1\,\text{\AA}$, grouping symmetry-related atoms of the same element into individual orbits. We then order the orbit blocks first by increasing Pauling electronegativity of the element and then by Wyckoff letter, while keeping atoms within each orbit contiguous.

During training, we apply a hierarchical orbit permutation to avoid dependence on an arbitrary atom enumeration. At the inter-orbit level, orbit blocks with the same element and the same Wyckoff label, including multiplicity, are randomly permuted without merging distinct orbits. At the intra-orbit level, atoms within each orbit block are independently permuted. The same permutation is applied jointly to atom types and fractional coordinates, while padding positions remain unchanged. Thus, the procedure changes only the ordering of atomic tokens and leaves the underlying crystal structure and PXRD pattern invariant. These random permutations are applied during training and disabled during validation and testing.

Together, this ordering and permutation scheme provides a consistent element- and symmetry-aware organization while avoiding dependence on any single equivalent atom enumeration. The element-set model further applies sinusoidal atom-slot positional encodings after ordering and permutation. These encodings specify atomic token positions and are distinct from the Fourier encoding of fractional atomic coordinates.

\textbf{Output heads.}
The backbone predicts atom-type, coordinate, and lattice outputs using
separate bias-enabled linear heads. The atom-type and coordinate heads
operate on each atomic token and map $512\rightarrow16$ and
$512\rightarrow3$, respectively, while the lattice head maps the lattice
token from $512\rightarrow6$ to predict the lower-triangular lattice latent.
No additional activation is applied after these heads.

Their raw outputs are converted to denoised estimates using EDM
preconditioning,
\begin{equation}
D_\theta(x,\sigma)
=
c_{\mathrm{skip}}(\sigma)x
+
c_{\mathrm{out}}(\sigma)F_\theta(\cdot),
\end{equation}
where
\begin{equation}
c_{\mathrm{skip}}(\sigma)
=
\frac{\sigma_{\mathrm{data}}^2}
     {\sigma^2+\sigma_{\mathrm{data}}^2},
\qquad
c_{\mathrm{out}}(\sigma)
=
\frac{\sigma\,\sigma_{\mathrm{data}}}
     {\sqrt{\sigma^2+\sigma_{\mathrm{data}}^2}}.
\end{equation}
Here, $\sigma_{\mathrm{data}}$ denotes the data-scale parameter used in
EDM preconditioning. Thus, the model heads predict continuous denoising
outputs rather than categorical atom-type logits or coordinate velocities.

\Cref{tab:model-configurations} summarizes the model configurations.

\begin{table}[!h]
\centering
\caption{
\textbf{\model{} model configurations.}
All configurations are shared between the known-composition and element-set
models unless specified otherwise.
Parameter counts include the PXRD encoder and conditioning modules.
}
\label{tab:model-configurations}
\begin{tabular}{lcc}
\toprule
\textbf{Component}
& \textbf{Known composition}
& \textbf{Element set $+N$} \\
\midrule

\multicolumn{3}{l}{\textbf{Backbone}} \\
\quad Hidden dimension
& \multicolumn{2}{c}{512} \\
\quad Transformer layers
& \multicolumn{2}{c}{14} \\
\quad Attention heads
& \multicolumn{2}{c}{16} \\
\quad FFN expansion ratio
& \multicolumn{2}{c}{4} \\
\quad FFN activation
& \multicolumn{2}{c}{GELU} \\
\quad Normalization
& \multicolumn{2}{c}{Pre-LN} \\
\quad LayerNorm affine / $\epsilon$
& \multicolumn{2}{c}{No / $10^{-6}$} \\
\quad Attention / FFN dropout
& \multicolumn{2}{c}{0 / 0} \\
\quad Final normalization
& \multicolumn{2}{c}{Affine-free LN} \\
\quad Time conditioning
& \multicolumn{2}{c}{adaLN-Zero} \\
\quad PXRD conditioning
& \multicolumn{2}{c}{In-context} \\
\quad Coordinate encoding
& \multicolumn{2}{c}{Fourier ($k_{\max}=32$)} \\

\midrule
\multicolumn{3}{l}{\textbf{Noise embedding}} \\
\quad Input
& \multicolumn{2}{c}{$\frac{1}{4}\log\sigma$} \\
\quad Sinusoidal frequencies / dimension
& \multicolumn{2}{c}{128 / 256} \\
\quad MLP
& \multicolumn{2}{c}{$512\!\rightarrow\!512\!\rightarrow\!512$} \\
\quad Activation
& \multicolumn{2}{c}{SiLU} \\

\midrule
\multicolumn{3}{l}{\textbf{Geometry Enhancement Module}} \\
\quad Sharing across layers
& \multicolumn{2}{c}{Shared} \\
\quad Distance / edge bias
& \multicolumn{2}{c}{Yes} \\
\quad Noise gate
& \multicolumn{2}{c}{Yes} \\

\midrule
\multicolumn{3}{l}{\textbf{PXRD encoder}} \\
\quad Hidden dimension
& \multicolumn{2}{c}{256} \\
\quad Transformer layers
& \multicolumn{2}{c}{4} \\
\quad Attention heads
& \multicolumn{2}{c}{8} \\
\quad Pooling
& \multicolumn{2}{c}{\texttt{[CLS]}} \\
\quad Dropout
& \multicolumn{2}{c}{0} \\
\quad Peak position encoding
& \multicolumn{2}{c}{Fourier ($F_{\max}=16$, $d=32$)} \\

\midrule
\multicolumn{3}{l}{\textbf{Output heads}} \\
\quad Coordinate
& \multicolumn{2}{c}{$512\!\rightarrow\!3$} \\
\quad Lattice
& \multicolumn{2}{c}{$512\!\rightarrow\!6$} \\
\quad Head activation
& \multicolumn{2}{c}{None} \\

\midrule
\multicolumn{3}{l}{\textbf{Setting-specific configuration}} \\
\quad Atom-type head
& Unused
& $512\!\rightarrow\!16$ \\
\quad Atom ordering
& None
& MCFlow \\
\quad Atom-slot positional encoding
& None
& Sinusoidal \\
\quad Parameters
& 71.01M
& 71.32M \\

\bottomrule
\end{tabular}
\end{table}

\clearpage
\subsection{Training objective and optimization}
\label{sec:training-configuration}

\textbf{Training objective.}
All \model{} models are trained with the EDM denoising objective used by the Crystalite backbone. We combine the atom-type, fractional-coordinate, and lattice losses as
\begin{equation}
    \mathcal{L}
    =
    \lambda_H \mathcal{L}_H
    +
    \lambda_F \mathcal{L}_F
    +
    \lambda_L \mathcal{L}_L,
\end{equation}
where $\mathcal{L}_H$, $\mathcal{L}_F$, and $\mathcal{L}_L$ denote the atom-type, fractional-coordinate, and lattice denoising losses, respectively. For the known-composition model, atom types remain fixed throughout diffusion, so the atom-type loss is inactive and we use $(\lambda_H,\lambda_F,\lambda_L)=(0,50,5)$. For the element-set model, atom types are jointly denoised with coordinates and lattice parameters, and we use $(64,150,5)$.

The two settings also use different EDM noise distributions. For known composition, we sample
\begin{equation}
    \log \sigma \sim \mathcal{N}(-1.2,1.2^2),
\end{equation}
with $(\sigma_{\min},\sigma_{\max})=(0.002,80)$ for coordinates and lattice variables; atom types remain fixed and are not noised. For element-set conditioning, we instead use
\begin{equation}
    \log \sigma \sim \mathcal{N}(-1.2,1.5^2),
\end{equation}
with $(\sigma_{\min},\sigma_{\max})=(0.001,40)$ for atom types, coordinates, and lattice variables. \Cref{tab:diffusion-configurations} summarizes these objective settings.

\begin{table}[!h]
\centering
\caption{
\textbf{Diffusion-objective settings.}
Loss weights and EDM noise-distribution parameters for the two chemical-input settings.
}
\label{tab:diffusion-configurations}
\begin{tabular}{lcc}
\toprule
& \textbf{Known composition} & \textbf{Element set $+N$} \\
\midrule
Loss weights $(\lambda_H,\lambda_F,\lambda_L)$
& $(0,50,5)$
& $(64,150,5)$ \\
EDM $(P_{\mathrm{mean}},P_{\mathrm{std}})$
& $(-1.2,1.2)$
& $(-1.2,1.5)$ \\
EDM $(\sigma_{\min},\sigma_{\max})$
& $(0.002,80)$
& $(0.001,40)$ \\
\bottomrule
\end{tabular}
\end{table}

\textbf{Optimization.}
We train all models with AdamW~\citep{loshchilov2019adamw} using $(\beta_1,\beta_2)=(0.9,0.999)$ and zero weight decay. We clip the global gradient norm at 1.0, use bfloat16 precision, and maintain an exponential moving average (EMA) of the model parameters with decay $0.9999$. Learning-rate settings depend on the chemical-input setting. The known-composition model uses a learning rate of $2\times10^{-4}$ with 1,000 linear warmup steps followed by a constant learning rate, whereas the element-set model uses a learning rate of $1\times10^{-4}$ with 500 warmup steps followed by cosine decay. \Cref{tab:optimization-conditioning} summarizes these conditioning-specific settings.

Training budgets depend on the training corpus. Models trained on MP-20, which contains structures with at most 20 atoms, are trained for 800K optimizer steps with a batch size of 512. Models trained on \dataset{}, which allows up to 52 atoms, are trained for 4M optimizer steps with a batch size of 256. These budgets correspond to approximately 15K and 250 epochs, respectively. All runs use a single NVIDIA RTX 3090 GPU without gradient accumulation and a training seed of 123. \Cref{tab:optimization-corpus} summarizes these corpus-specific settings.

\begin{table}[!h]
\centering
\caption{
\textbf{Conditioning-specific optimization settings.}
}
\label{tab:optimization-conditioning}
\begin{tabular}{lcc}
\toprule
& \textbf{Known composition} & \textbf{Element set $+N$} \\
\midrule
Learning rate
& $2\times10^{-4}$ & $1\times10^{-4}$ \\
LR schedule
& Constant & Cosine decay \\
Warmup steps
& 1,000 & 500 \\
\bottomrule
\end{tabular}
\end{table}

\begin{table}[!h]
\centering
\caption{
\textbf{Corpus-specific training settings.}
}
\label{tab:optimization-corpus}
\begin{tabular}{lcc}
\toprule
& \textbf{MP-20} & \textbf{PXRD-Bench} \\
\midrule
Maximum atom count $N_{\max}$
& 20 & 52 \\
Training steps
& 800K & 4M \\
Approx.\ epochs
& $\sim$15K & $\sim$250 \\
Batch size
& 512 & 256 \\
GPUs / gradient accumulation
& 1 / 1 & 1 / 1 \\
Training seed
& 123 & 123 \\
\bottomrule
\end{tabular}
\end{table}

\textbf{Ablation training settings.}
Unless otherwise specified, the ablation experiments use the same
MP-20 architecture, training objective, and optimization settings
described above. Each ablation model is trained from scratch for
200K optimizer steps, corresponding to approximately 3,846 epochs,
and comparisons use the final 200K-step EMA checkpoint rather than
validation-based checkpoint selection.

For the augmentation ablation, we train three independent models with
training seeds $\{123,124,125\}$. Each trained model is evaluated using
five sampling seeds $\{777,778,779,780,781\}$. We first average over
the five sampling seeds for each trained model and then report the mean
and standard deviation across the three training seeds.

For the peak-representation and encoding studies, we instead train a single model with training seed $123$
and evaluate it using the same five sampling seeds.
For the representation and encoding studies, we keep the full
augmentation setting fixed and vary only the component under study.
For the augmentation ablation, we vary the specified augmentation
components while keeping the remaining training configuration fixed.
\Cref{tab:ablation-training-configurations} summarizes the settings
that differ from the main training protocol.

\begin{table}[t]
\centering
\caption{
\textbf{Ablation training settings.}
Settings that differ from the main training protocol.
}
\label{tab:ablation-training-configurations}
\begin{tabular}{ll}
\toprule
\textbf{Setting} & \textbf{Value} \\
\midrule
Training steps & 200K ($\sim$3,846 epochs) \\
Checkpoint & Final 200K-step EMA checkpoint \\
Training seed (representation)
& $123$ \\
Training seeds (augmentation)
& $\{123,124,125\}$ \\
Sampling seeds
& $\{777,778,779,780,781\}$ per training run \\
\bottomrule
\end{tabular}
\end{table}

\textbf{Multiphase fine-tuning.}
We initialize the phase-wise structure generator from the pretrained
element-set Bragglite checkpoint and fine-tune it on the simulated
multiphase training set. We use AdamW with a learning rate of
$1\times10^{-5}$, 500 linear warmup steps followed by cosine decay,
and a batch size of 128 phase examples. We fine-tune the model for
50K optimizer steps using a training seed of 123.

XDecomposer is initialized from its publicly released checkpoint and
fine-tuned separately on the simulated multiphase training set. We use
AdamW with an initial learning rate of $3\times10^{-6}$, reducing the
learning rate by a factor of two when the validation metric plateaus.
We use a batch size of 16 mixtures and fine-tune for 100K optimizer
steps, corresponding to approximately 21 epochs, with a training seed
of 123.

For both models, we use the final fine-tuned checkpoint for evaluation.

\subsection{Sampling and checkpoint selection}
\label{appx:sampling-checkpoint}

\textbf{Sampling.}
At inference time, we sample from the EMA model using the stochastic Heun sampler from EDM~\citep{karras2022elucidating}. For each input PXRD pattern, we run 150 sampling steps and generate 20 candidate structures. The same sampling procedure is used for models trained on MP-20 and PXRD-Bench. We follow the sampling configuration of Crystalite~\citep{veljkovic2026crystalite}, including the Karras noise schedule, stochastic churn, and anti-annealing settings. \Cref{tab:sampling-configurations} summarizes the sampler configuration.

\begin{table}[!h]
\centering
\caption{
\textbf{Sampling configurations.}
Settings are shared across MP-20 and PXRD-Bench and follow the Crystalite sampling configuration.
}
\label{tab:sampling-configurations}
\begin{tabular}{lcc}
\toprule
\textbf{Setting}
& \textbf{Known composition}
& \textbf{Element set $+N$} \\
\midrule
Solver / steps
& \multicolumn{2}{c}{Heun / 150} \\
Noise schedule
& \multicolumn{2}{c}{Karras, $\rho=7$} \\
$S_{\mathrm{churn}}$
& \multicolumn{2}{c}{60} \\
Churn range $[S_{\min},S_{\max}]$
& \multicolumn{2}{c}{$[0,999]$} \\
$S_{\mathrm{noise}}$
& \multicolumn{2}{c}{1.003} \\
$\sigma_{\mathrm{data}}$
& \multicolumn{2}{c}{0.3} \\
Anti-annealing $(H,F,L)$
& $(0,10,10)$
& $(4,4,4)$ \\
\bottomrule
\end{tabular}
\end{table}

\textbf{Checkpoint selection.}
We select checkpoints using a fixed random subset of 512 structures from the simulated PXRD validation set. Every 10K training steps, we evaluate one generated candidate per validation structure and compute the top-1 structure match rate. We report the EMA weights from the checkpoint with the highest validation top-1 match rate rather than automatically using the final training checkpoint.

\subsection{Baseline configurations}
\label{appx:baseline-configurations}
We use the publicly released implementation of each baseline rather
than reimplementing the models. Across training corpora, all baselines
are trained using Adam with $(\beta_1,\beta_2)=(0.9,0.999)$,
$\epsilon=10^{-8}$, and no weight decay. We use a training seed of 42,
a single NVIDIA GeForce RTX 3090 GPU, no gradient accumulation, and
elementwise gradient clipping at a maximum absolute value of 0.5.
For all baselines, we maintain exponential moving average (EMA)
parameters and use EMA checkpoints for evaluation.

\textbf{PXRD-Bench training.}
For baselines trained on PXRD-Bench, we use a batch size of 256 and
initialize the learning rate to $10^{-3}$. We use
\texttt{ReduceLROnPlateau}, monitoring the Top-1 structure match rate
on a fixed subset of 512 structures from the simulated validation set.
The learning rate is reduced by a factor of two if this metric does not
improve for five epochs, and training is terminated when the learning
rate would fall below $10^{-5}$. We select the EMA checkpoint with the
highest monitored validation Top-1 match rate for evaluation.

\textbf{MP-20 training.}
For baselines trained on MP-20 or MP-20-polymorph-split, we use the
same optimization protocol but a longer learning-rate patience and a
fixed maximum training budget. Starting from a learning rate of
$10^{-3}$, we use \texttt{ReduceLROnPlateau} with a patience of
30 epochs and a minimum learning rate of $10^{-5}$, again monitoring
the Top-1 structure match rate on a fixed subset of 512 structures
from the corresponding simulated validation set. Training is run for
at most 300K optimizer steps, and we use the EMA checkpoint with the
highest monitored validation Top-1 match rate for evaluation.

\textbf{DiffCSP.}
DiffCSP~\citep{jiao2023diffcsp} serves as a composition-only CSP
baseline and does not receive PXRD input.
We condition the model on the exact primitive-cell composition and use
a six-layer CSPNet with hidden dimension 512 and a time-embedding
dimension of 256.
The diffusion process uses 1,000 steps.

\textbf{XRDiff.}
For XRDiff~\citep{segal2026xrdiff}, the PXRD profile is represented on
850 bins over $2\theta\in[5^\circ,90^\circ]$.
We retain up to the 20 strongest detected peaks and represent each peak
using its intensity, position, prominence, and width, yielding an
80-dimensional fixed-size PXRD feature.
The structure generator uses a six-layer CSPNet with hidden dimension
512 and FiLM conditioning on the PXRD representation.
We do not apply additional stochastic PXRD augmentation during
training.
We train both the known-composition and element-set $+N$ variants;
for the latter, the atom types are additionally predicted and included
in the training objective.

\textbf{RealPXRD-Solver.}
RealPXRD-Solver~\citep{li2026realpxrdsolver} uses a 750-bin profile over
$2\theta\in[5^\circ,80^\circ]$, from which a Bragg-stick
representation is constructed.
We use its two-layer Transformer PXRD encoder together with a
six-layer CSPNet of hidden dimension 512.
For simulated-to-experimental transfer, we follow the PXRD
augmentation procedure described in the original method.

\textbf{PXRDGen.}
PXRDGen~\citep{li2025pxrdgen} receives a 7,500-bin Bragg-derived PXRD
profile over $2\theta\in[5^\circ,80^\circ]$.
We first contrastively pretrain its PXRD CNN encoder on the
PXRD-Bench training corpus and subsequently train the diffusion model
while continuing to update the CNN jointly.
We use a contrastive temperature of 1.0 and do not use an externally
pretrained PXRD encoder.

\textbf{Crystalyze.}
Crystalyze~\citep{riesel2024crystalyze} operates on an 8,500-bin PXRD
profile using its PXRD CNN encoder, a latent dimension of 256, and a
GemNet-T structure decoder.
During training, we sample one of the four provided Caglioti
broadening settings and add the publicly released RRUFF-derived Gamma
background-noise augmentation.
We do not apply an additional peak-intensity modulation.

\textbf{Baseline sampling.}
For each baseline, we use the default sampling procedure provided by
its public implementation.
Crystalyze uses 5,000-step annealed Langevin dynamics.
DiffCSP, XRDiff, and PXRDGen use 1,000 predictor--corrector sampling
steps, corresponding to 2,000 neural network evaluations.
RealPXRD-Solver uses a 200-step Euler integrator.

\subsection{Compute resources}
\label{appx:compute-resources}

All \model{} models are trained on a single NVIDIA GeForce RTX 3090 GPU with 24\,GB of memory. \Cref{tab:training-cost} reports the training configuration and computational cost of the main models, including batch size, number of optimization steps, wall-clock training time, and peak GPU memory usage.

\begin{table}[h]
\caption{
\textbf{Training cost.}
Wall-clock training time and peak GPU memory usage for the main \model{} configurations. All models are trained on a single NVIDIA GeForce RTX 3090 GPU.
}
\label{tab:training-cost}
\centering
\begingroup
\small
\setlength{\tabcolsep}{5pt}
\renewcommand{\arraystretch}{1.12}
\begin{tabular}{lcccc}
\toprule
\textbf{Training data / input}
& \textbf{Batch size}
& \textbf{Steps}
& \textbf{Time (h)}
& \textbf{Peak memory (GiB)} \\
\midrule
\dataset{} / Known composition
& 256 & 4M & 200.18 & 7.97 \\
\dataset{} / Element set
& 256 & 4M & 201.29 & 7.98 \\
MP-20 / Known composition
& 512 & 800K & 37.10 & 7.15 \\
MP-20 / Element set
& 512 & 800K & 37.89 & 7.16 \\
\bottomrule
\end{tabular}
\endgroup
\end{table}
\newpage
\section{Qualitative examples}

\begin{figure*}[h]
    \centering
    \includegraphics[width=\textwidth]{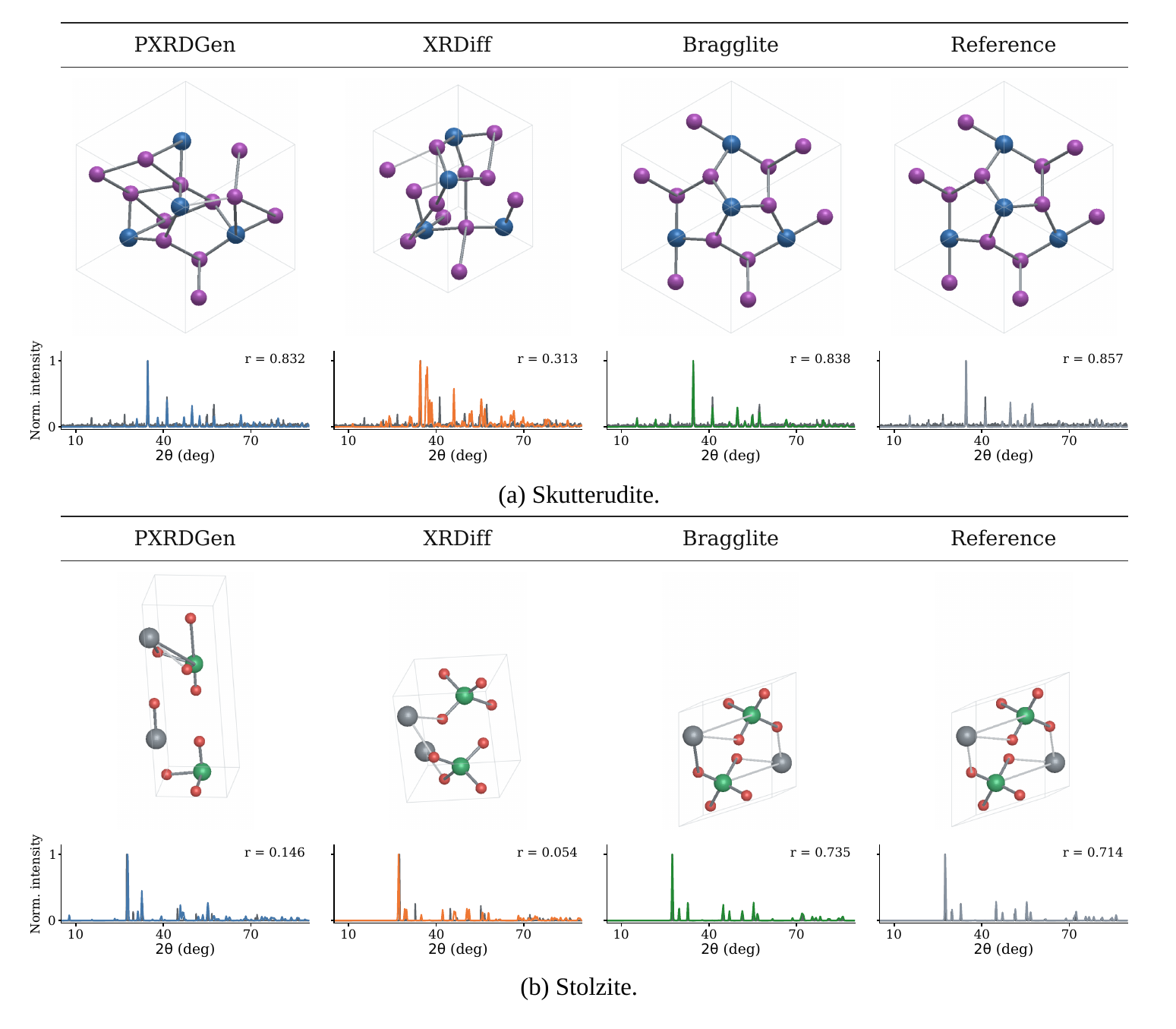}
    \label{fig:qualitative-stolzite}
\caption{
\textbf{Qualitative comparison of structures} generated by PXRDGen, XRDiff, and \model{} on experimental PXRD patterns, using Skutterudite and Stolzite examples from the RRUFF test split. For each method, the generated crystal structure and its simulated PXRD pattern are shown alongside the reference structure and experimental pattern. The reported $r$ denotes the correlation between the simulated and experimental PXRD profiles.
}
    \label{fig:qualitative-examples}
\end{figure*}

We provide visualizations of representative predictions from the evaluated models in \cref{fig:qualitative-examples}.

\newpage
\section{Scope and limitations}
\label{appx:limitations}

\textbf{Candidate generation vs. end-to-end determination.}
Our benchmark focuses on the candidate-generation stage of PXRD-based structure determination rather than complete end-to-end crystallographic analysis. We evaluate generated structures before downstream refinement and use Top-$K$ structural match rate as a direct and reproducible measure of structure recovery. This enables controlled comparison of generative models without confounding differences in refinement or other post-processing procedures. However, structural match rate is not equivalent to end-to-end structure-determination success: unmatched candidates may still be refinable to the correct structure, while matched candidates do not necessarily guarantee successful refinement against the measured pattern. Developing standardized automated refinement and validation procedures with minimal expert intervention would enable more complete end-to-end evaluation and is an important direction for future work.

\textbf{Oracle atom counts.}
A limitation of the current benchmark is that the primitive-cell atom count is provided as oracle information. As in many diffusion-based crystal structure prediction and generation models, the number of atomic sites is specified before structure generation. Thus, our element-set setting requires the model to infer elemental multiplicities but not the total atom count, while the two-phase setting provides the atom count of each constituent phase. A more general formulation would infer this quantity from the chemical and diffraction information, for example by learning an atom-count distribution or predicting the count jointly with the crystal structure.

\textbf{Limited experimental PXRD data.}
A further limitation is the availability of experimental PXRD data paired with reliable structural labels. Publicly available measured PXRD patterns are limited, particularly when a corresponding crystal structure is required as a trustworthy reference for evaluation or training. This constrains both the scale and coverage of experimental benchmarks and the ability to study adaptation directly on measured diffraction data. Building larger experimental PXRD datasets with reliable and traceable structure annotations would enable broader evaluation across materials and measurement conditions, as well as systematic study of experimental fine-tuning and adaptation.

\end{document}